\documentclass[lettersize,journal]{IEEEtran}
\usepackage{stfloats}
\usepackage{float}
\usepackage{url}
\usepackage{verbatim}
\usepackage{graphicx}
\usepackage{cite}
\usepackage{caption}
\usepackage{subcaption}
\usepackage{tikz}
\usepackage{amsmath,amssymb,amsfonts}
\usepackage{textcomp}

\usepackage[export]{adjustbox}
\usepackage{pgffor}

\usepackage{multirow}
\usepackage[hidelinks]{hyperref}
\usepackage{makecell}
\usepackage{picinpar}
\usepackage[latin1]{inputenc}
\usepackage{colortbl}
\usepackage{soul}
\usepackage{pifont}
\usepackage{color}
\usepackage{alltt}
\usepackage{enumerate}
\usepackage{siunitx}
\usepackage{breakurl}
\usepackage{epstopdf}
\usepackage{pbox}
\usepackage{gensymb}
\usepackage{makecell}
\usepackage{bm}
\usepackage[misc]{ifsym}
\usepackage{balance}
\usepackage{booktabs}
\usepackage{enumitem} 
\usepackage{etoolbox}
\usepackage{algpseudocode}
\usepackage{algorithm}
\usepackage{lipsum} 

\algrenewcommand\algorithmicrequire{\textbf{Input:}}
\algrenewcommand\algorithmicensure{\textbf{Output:}}

\newcommand{\AlgPhase}[1]{%
    \vspace{0.5em}
    \Statex\hspace{-\algorithmicindent}\textbf{#1}
    \vspace{0.5em}
}

\algrenewcommand\algorithmiccomment[1]{\hfill\textit{// #1}}

\usetikzlibrary{positioning, arrows.meta}
\tikzset{
    block/.style = {rectangle, draw, text centered, minimum height=2em, minimum width=12em},
    arrow/.style = {thick,->,>=Stealth},
    splitarrow/.style = {thick,->,>=Stealth, shorten >=2pt, shorten <=2pt}
}

\begin{document}
\title{\LARGE \bf
Automated Action Generation based on Action Field for Robotic Garment Smoothing and Alignment}

\author{Hu Cheng,~\IEEEmembership{Member,~IEEE,}
        Fuyuki Tokuda,~\IEEEmembership{Member,~IEEE,}
        Kazuhiro Kosuge,~\IEEEmembership{Life Fellow,~IEEE}

\thanks{This work was supported in part by the Innovation and Technology Commission of the HKSAR Government through InnoHK initiative, in part by the JC STEM Lab of Robotics for Soft Materials through The Hong Kong Jockey Club Charities Trust, and in part by Tohoku University through the Joint Research Program. \textit{(Corresponding author: Hu Cheng.)
}}

\thanks{Hu Cheng, Fuyuki Tokuda, and Kazuhiro Kosuge are with the JC STEM Lab of Robotics for Soft Materials, Department of Electrical and Electronic Engineering, Faculty of Engineering, The University of Hong Kong, Hong Kong SAR, China (email: hucheng@hku.hk, fuyuki.tokuda.b3@tohoku.ac.jp, kosuge@hku.hk)}

}

\maketitle
\begin{abstract}
Garment manipulation using robotic systems is a challenging task due to the diverse shapes and deformable nature of fabric. In this paper, we propose a novel method for robotic garment smoothing and alignment that significantly improves the accuracy while reducing computational time compared to previous approaches. Our method features an action generator that directly interprets scene images and generates pixel-wise end-effector action vectors using a neural network. The network also predicts a manipulation score map that ranks potential actions, allowing the system to select the most effective action. Extensive simulation experiments demonstrate that our method achieves higher smoothing and alignment performances and faster computation time than previous approaches. Real-world experiments show that the proposed method generalizes well to different garment types and successfully flattens garments. 
\end{abstract}

\begin{mynote}
Vision-based robotic garment manipulation faces significant complexities due to garments' diverse shapes and high-dimensional states, which pose challenges for both state perception and action generation. In this paper, we propose a novel deep neural network that can generate actions to smooth various garments from their RGB images. Compared to existing methods, our method generates pixel-level actions across the entire garment area, each providing a predicted manipulation score that assists in the selection of a final manipulation action. In addition, the generation requires only a single-shot network forward computation, which significantly improves efficiency. The training data consists of large-scale recorded garment state parameters and the corresponding manipulating actions in the simulator. Real-world experiments demonstrate the effectiveness and generalization capability of our model. 
\end{mynote}


\section{Introduction}
\IEEEPARstart{R}{obotic} manipulation of deformable objects is essential in various applications, such as cable routing\cite{luo2024multi}, bag opening\cite{chen2023autobag}, and garment manipulation\cite{li2018model, huang2025sis, 10966003}. Among these, garment and fabric manipulation present unique challenges due to their high degrees of freedom, self-occlusion, and complex nonlinear material properties. These characteristics make it difficult to estimate the state of the fabric and generate actions for manipulators. Vision-based deep learning methods have been explored to address these challenges. In this paper, we propose an action generation model for a garment manipulation system, focusing on the tasks of smoothing and aligning the garment, which are critical prerequisites for the following procedures, such as defect inspection, printing, and embroidery. Currently, the action generation methods can be generally categorized into two groups: one is the two-stage method that first estimates the fabric state and then heuristically generates actions based on the state; the other one is the single-stage method, which directly outputs action from the visual input. 

The performance of two-stage methods depends heavily on the accuracy of fabric state estimation, which is often computationally expensive and sensitive to self-occlusion. Several techniques, such as dense visual correspondences by Sundaresan \textit{et al.} \cite{sundaresan2020learning}, garment mesh modeling by Chi and Song \cite{chi2021garmentnets}, and semantic keypoint extraction by Deng and Hsu \cite{deng2025general}, have been introduced to improve garment state estimation. While these methods have significantly advanced the accuracy and robustness of garment state estimation, challenges remain in handling complex deformations and occlusions. 

\begin{figure}[tbp]
  \vspace{0.32em}
  \centering
  \begin{subfigure}{0.43\columnwidth}
    \centering
    \includegraphics[height=1.5in]{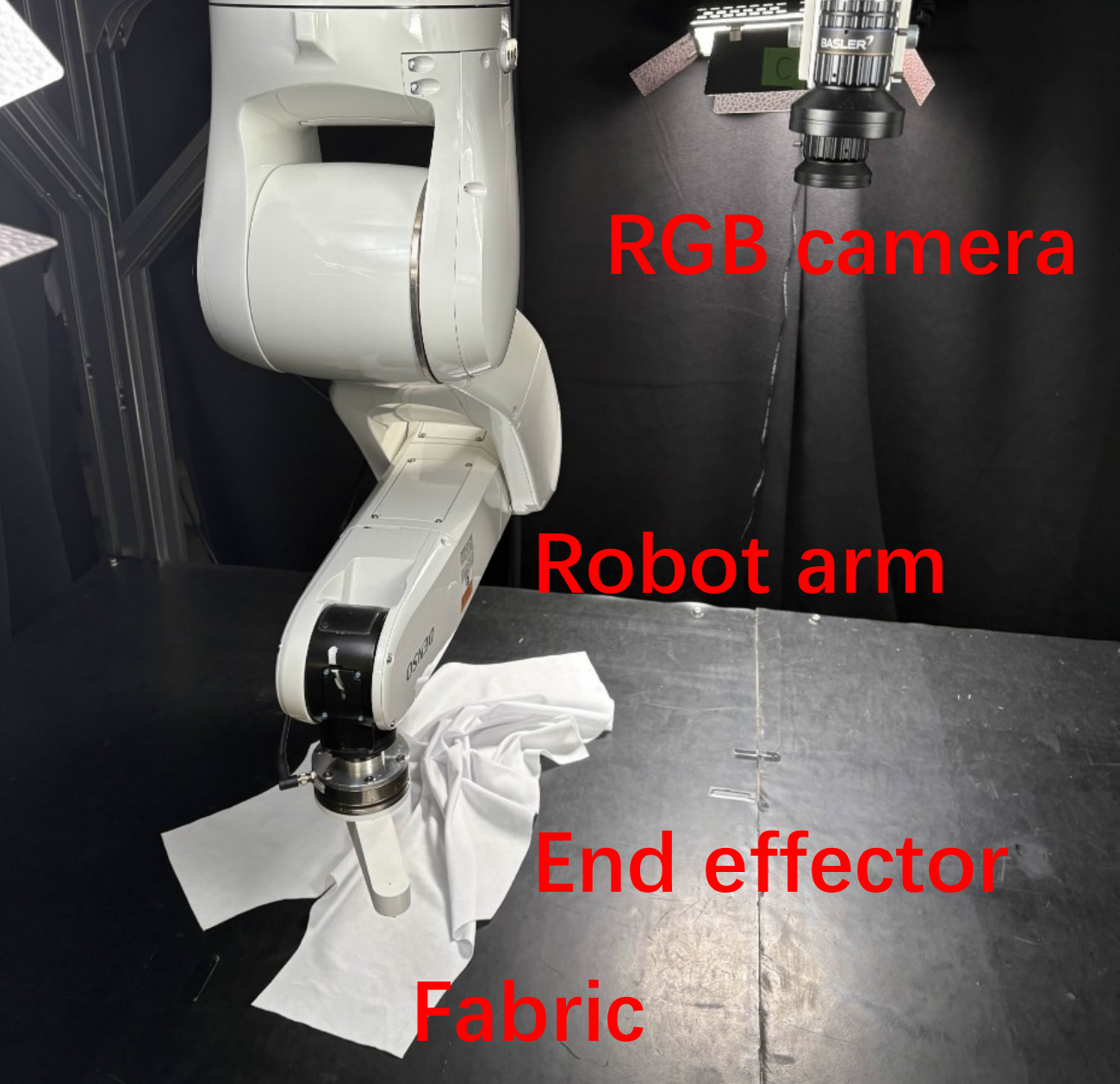}
    \caption{Robot Platform}
    \label{fig:robotplatform}
  \end{subfigure}%
  \hspace{0.1mm}
  \begin{subfigure}{0.46\columnwidth}
    \centering
    \includegraphics[height=1.5in]{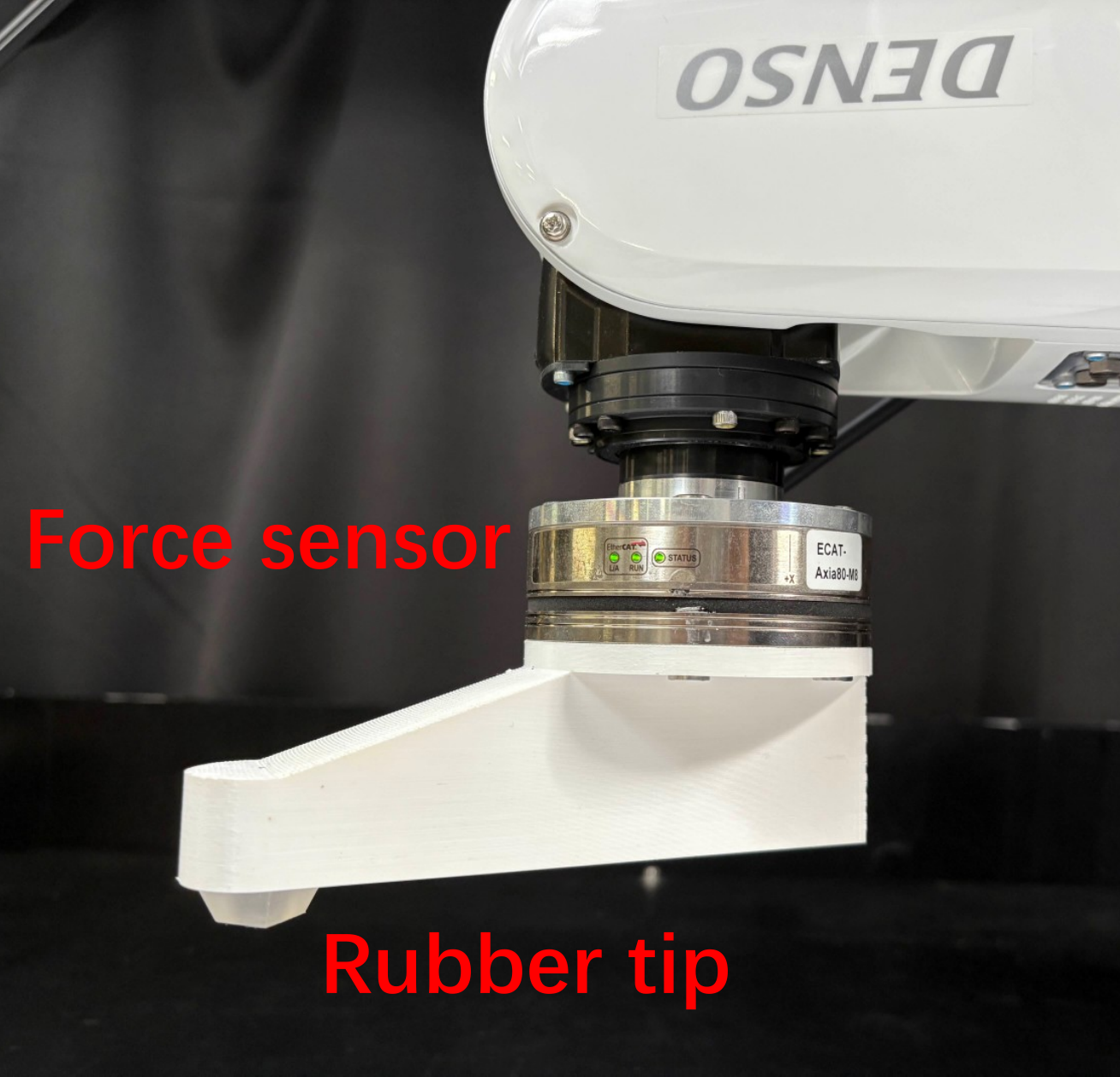}
    \caption{End-effector}
    \label{fig:end-effector}
  \end{subfigure}
  \caption{(a) is the robot platform and (b) shows the end-effector used to manipulate the fabric.}
  \label{fig:hardware}
\end{figure}

Single-stage methods \cite{ha2022flingbot, canberk2023cloth, yang2024clothppo, zhou2025dualarm} directly output the garment manipulation action from visual observations, which eliminates the need for explicit fabric state estimation and simplifies the action generation. These methods have proven effective in various manipulation tasks and have contributed to simplifying the action generation compared to the two-stage methods. However, many single-stage approaches rely on the spatial action map strategy \cite{zeng2018learning}, which involves multiple feed-forwards of the network and often represents flattening actions such as the pulling direction and pulling distance in predefined and discretized values. The multiple forward passes increase computational load and reduce efficiency. In addition, for these methods, the problem is reformulated into selecting manipulation points on a garment that match a finite set of predefined action configurations, i.e., [angle, distance] pairs. In other words, the predicted action direction and distance are coupled to the action position. These methods prevent the estimated actions from covering the entire garment, thereby reducing the flexibility in real-world manipulation. With these sparse and potentially clustered predicted actions, the algorithms will also find it hard to infer the overall effect of applying the actions during the forward process. This raises the central problem addressed in this work, i.e., decoupling the configurations of action from its position so that the robot can generate continuous, unconstrained actions on a garment, while keeping the prediction process in a single shot to maintain a high efficiency. To this end, we propose a reversed procedure. We first estimate the possibility of effective actions in each pixel location and then directly predict the corresponding angle and distance. This decoupling not only yields dense action estimations but also leads to supervising the action configurations with target shape information. 

The proposed system is designed to simulate the ``pull" action to slide a garment on a plane. This action does not require a large spatial space, and the end-effector also has a simple structure, which is both robust and cost-effective compared to parallel grippers, dexterous hands, and suction cups. This ``pull'' action has high execution success rate and relieves the challenges of thin object grasping \cite{cheng2022robot, 10132389, cheng2026vision}. Building on this system, we propose a novel learning-based action-generation framework that directly interprets scene images and generates the manipulation score, distance, and angle maps, simultaneously. These maps are then converted to pixel-wise end-effector action vectors, i.e., action field. By representing the manipulator's action as pixel-wise end-effector action vectors, our method only requires a single forward propagation of the network to generate an action of the manipulator. Our approach improves the accuracy and efficiency of garment manipulation compared to the previous methods. 

We propose a robotic garment and fabric manipulation system as shown in Fig.~\ref{fig:robotplatform}, which accomplishes the smoothing and positioning of different garments through planar actions by a low-cost and robust single-tip end-effector. The main contributions are summarized as follows: 
\begin{itemize}
    \item The proposed method consists of an action generation method that can directly generate pixel-wise action field from the image of a crumpled garment, significantly increasing efficiency compared to the existing single-stage garment manipulation method.
    \item  We claim that the score map training suffers from the class imbalance, and propose to incorporate the semantic information of the garment to address this issue. In addition, we leverage the action field representation and propose a novel training scheme named \textit{shape loss} to supervise the learning of the distance and angle map. 
    \item Extensive simulation experiments demonstrate that our method achieves a higher coverage index and alignment index and faster computation time than previous approaches. 
    \item Real-world experiments further validate the robustness of our approach, showing that the proposed method generalizes well to different garment types and successfully flattens garments.
\end{itemize}

In the following, Section II presents related work for the deformable object manipulation. Section III explains the action generator structure and training modules. Section IV presents the experimental results and ablation studies. Section V concludes this paper. 

\section{Related Work}
The vision-based deformable object manipulation policy depends on the state perception methods and the action generator strategy. In this section, we first introduce state representations used in the two-stage methods. Then, we introduce single-stage methods based on the spatial action map strategy. 

\subsection{State Representations in Two-Stage Methods}
\subsubsection{Feature-based state representation}
Feature-based state representation relies on visual correspondence establishment or direct key region detection. Specifically, visual correspondence-based methods \cite{schmidt2016self, florence2018dense, tokuda2025transformer} establish a mapping between image pixels representing the same surface point on an object. These correspondences capture the deformable object state, enabling the policy generation for the rope knot-tying \cite{sundaresan2020learning} or fabric smoothing and unfolding \cite{ganapathi2021learning, wu2024unigarmentmanip}. Meanwhile, \cite{berenson2013manipulation, wu2019learning, matas2018sim} calculated the sparse correspondence between the semantic keypoints to reduce the computational burden. Key region-based methods focus on detecting geometric \cite{GPTFabric2024} or semantic \cite{chen2023learning, chen2023autobag, clark2023household} regions on objects rather than individual keypoints, and then integrate them with a heuristic or model-based \cite{achiam2023gpt} high-level planner. However, these methods are hard to capture small wrinkles or subtle shape changes of the fabric. In addition, hand-crafted features and rule-based policies limit their ability to handle different types of garments. 

\subsubsection{3D reconstruction}
3D reconstruction provides complete geometric configurations of the fabric to represent its state. VCD \cite{lin2022learning} and TRTM \cite{wang2024trtm} modeled the fabric using Graph Neural Networks (GNNs), while Chen \textit{et al.} \cite{chi2021garmentnets} estimated a full 3D mesh in a canonical frame through mesh-completion. VCD \cite{lin2022learning} reconstructed the garment mesh with visual input and inferred the connectivity of the visible points. Meanwhile, Huang \textit{et al.} \cite{huang2022mesh} explicitly reasoned the occluded regions to achieve a full reconstructed mesh dynamics model. While these methods enable full state estimation with high computational overhead, the policy effectiveness is significantly affected by the reconstruction accuracy. 

\subsubsection{Optical-Flow-Like representation} 
Another type of two-stage framework uses optical-flow-like descriptors to unify state estimation and policy generation. Weng \textit{et al.} \cite{weng2022fabricflownet} learned cloth-folding actions by generating a flow map from the current and goal images to infer placement points, while an auxiliary network selects the pick point. Agarwal \textit{et al.} \cite{agarwal2023point} extended this concept by computing 3D point-cloud correspondences. The proposed shape loss employs a representation format similar to the flow map. However, it is based on fundamentally different objectives and is regarded as a training procedure. Specifically, the shape loss serves as a supervisory signal, evaluating how effectively the entire predicted dense actions deform the garment toward the target shape. This loss accelerates garment alignment in early manipulation stages and improves final accuracy.

\subsection{Single-Stage Methods with Spatial Action Map}
The single-stage methods avoid establishing descriptors by generating actions directly from raw sensory observations. The spatial action map was initially proposed to predict a single-channel reward map that indicates the most suitable manipulation point \cite{zeng2018learning} or moving target \cite{wu2020spatial}. This approach was later extended to infer more complex action configurations with pulling direction and distance \cite{ha2022flingbot, Xu-RSS-22}. By applying a discrete set of rotation and scaling primitives to the input, these methods generate input variants that are individually fed into the network to yield a stack of reward maps, with the map containing the maximum value selected to determine the action. This strategy has been widely adopted in garment manipulation \cite{canberk2023cloth, blanco2023qdp, yang2024clothppo} and plastic bag knotting tasks \cite{gao2023iterative}.

However, the spatial action map approach is computationally intensive, as it requires multiple forward passes of the network for each rotated and scaled input image. Furthermore, because action parameters such as pulling direction and distance are sampled from a discrete set, the selected action may not be globally optimal. In contrast, our method predicts pixel-wise action maps with continuous values for score, pulling direction, and distance in a single forward pass, enabling more efficient and precise action generation. 

\section{Methodology}
In this section, we first describe how we define the planar robot action that manipulates fabrics. Then, we elaborate on the structure of the action generation network and the corresponding design of the loss function. Finally, we present the details of the collected training data. 

\subsection{Action Representations}\label{sec:action-representations}
\begin{figure}[tbp]
  \vspace{0.32em}
  \centering
  \begin{subfigure}{0.45\columnwidth} %
    \centering
    \includegraphics[height=1.1in]{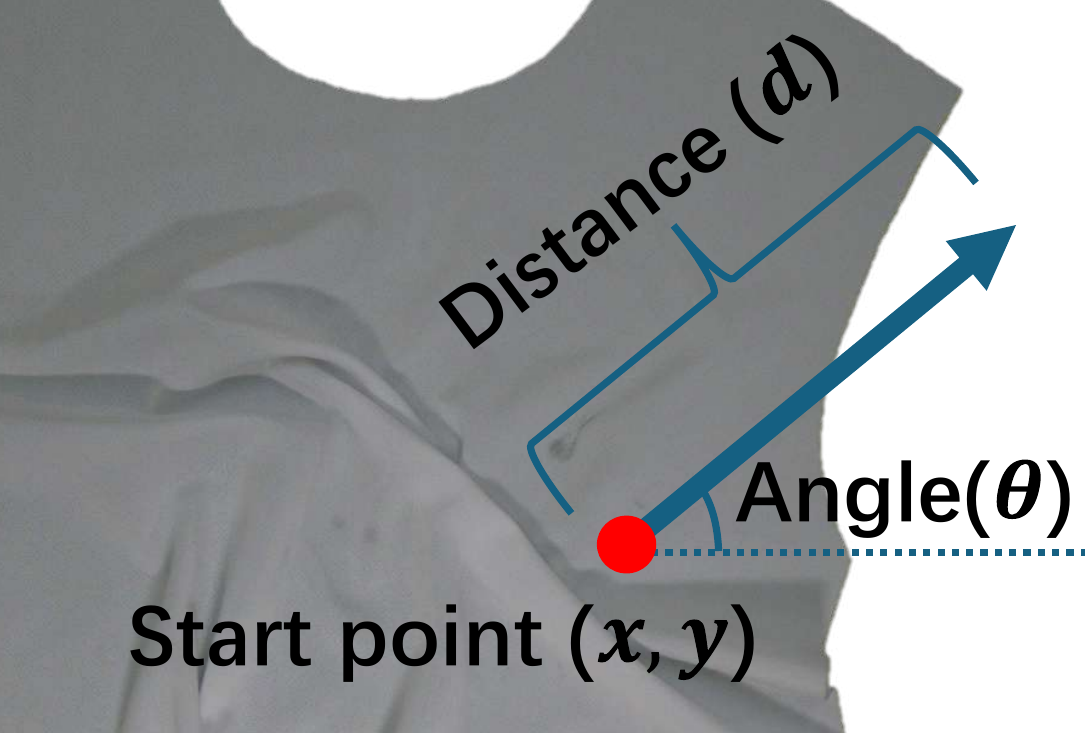}
    \caption{Action definition}
    \label{fig:action-definition}
  \end{subfigure}%
  \hspace{0.1mm}
  \begin{subfigure}{0.45\columnwidth} %
    \centering
    \includegraphics[height=1.1in]{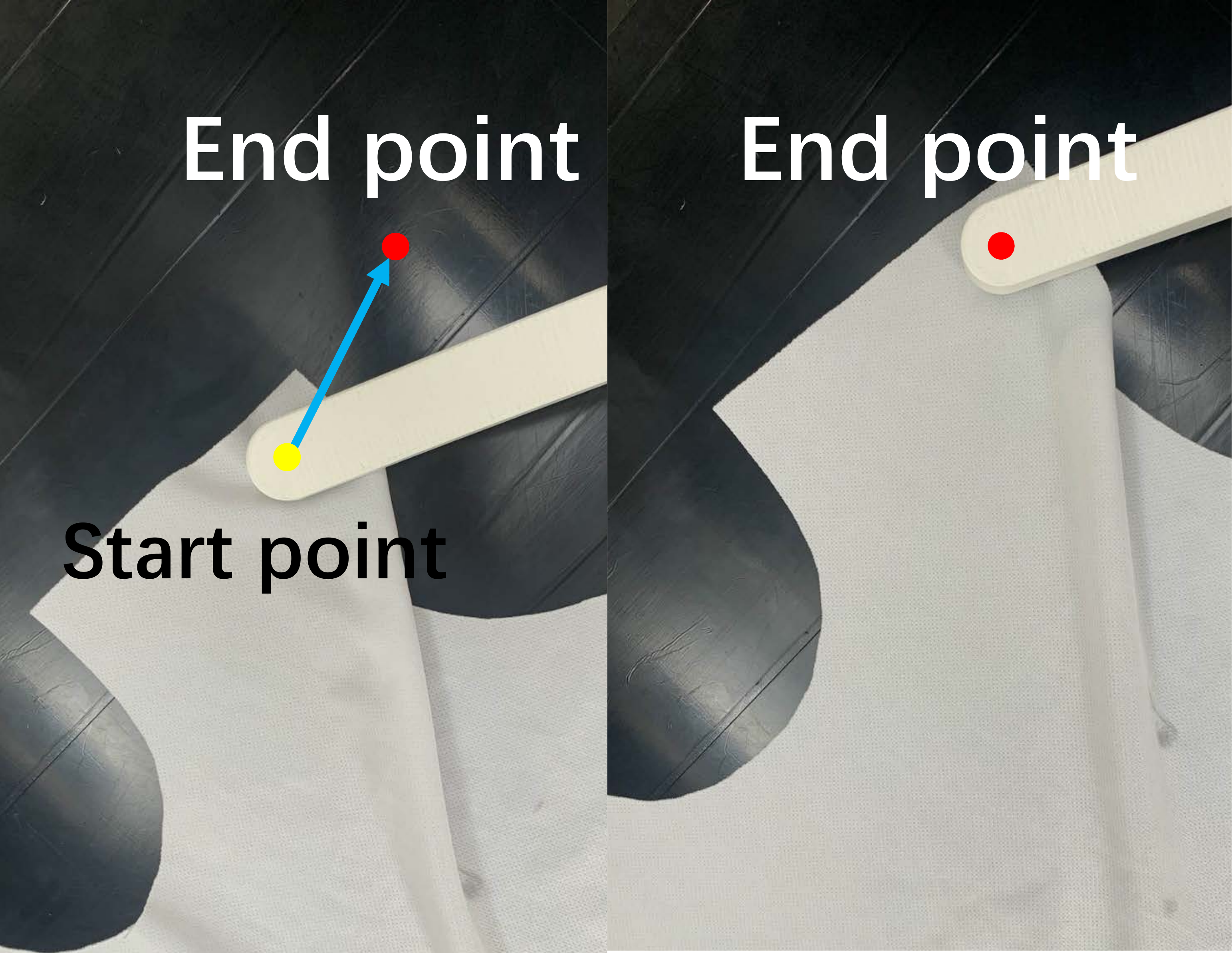}
    \caption{Action effect}
    \label{fig:action-effect}
  \end{subfigure}
  \caption{(a) The action that manipulates the fabric contains the start point $(x, y)$ in the image, the moving distance $d$, and the moving angle $\theta$. (b) demonstrates the sliding effect of the fabric by applying the action. }
\end{figure}

In this paper, we consider a task in which a robot manipulates a garment using planar sliding actions. The objective consists of two subtasks: (1) flattening the garment to eliminate wrinkles and increase its coverage area, and (2) aligning the garment to a predefined target pose with the desired position and orientation. Both subtasks are achieved through a sequence of planar actions executed by a simple single-tip end-effector. The end-effector with a rubber tip is attached to the manipulator as shown in Fig.~\ref{fig:end-effector}. The rubber tips exhibit high friction against the fabric, whereas the friction between the fabric and the table is relatively low. For this planar action, we define it using a 4D vector: 
$[x, y, \theta, d]$, as shown in Fig.~\ref{fig:action-definition}. The starting point $(x, y)$ is the initial contact position in image space where the end-effector first presses down the fabric. A downward force is then applied to press the fabric firmly against the table, ensuring stable contact during the action. Finally, the robot arm moves along a straight path in the direction of angle $\theta$ with a distance of $d$. This action causes the fabric to slide on the table surface and smooth the fabric, as shown in Fig.~\ref{fig:action-effect}.

\subsection{Action Generator}

\begin{figure*}[tbp] 
\vspace{-0.2cm}
 \centering
 \includegraphics[width=0.9\textwidth]{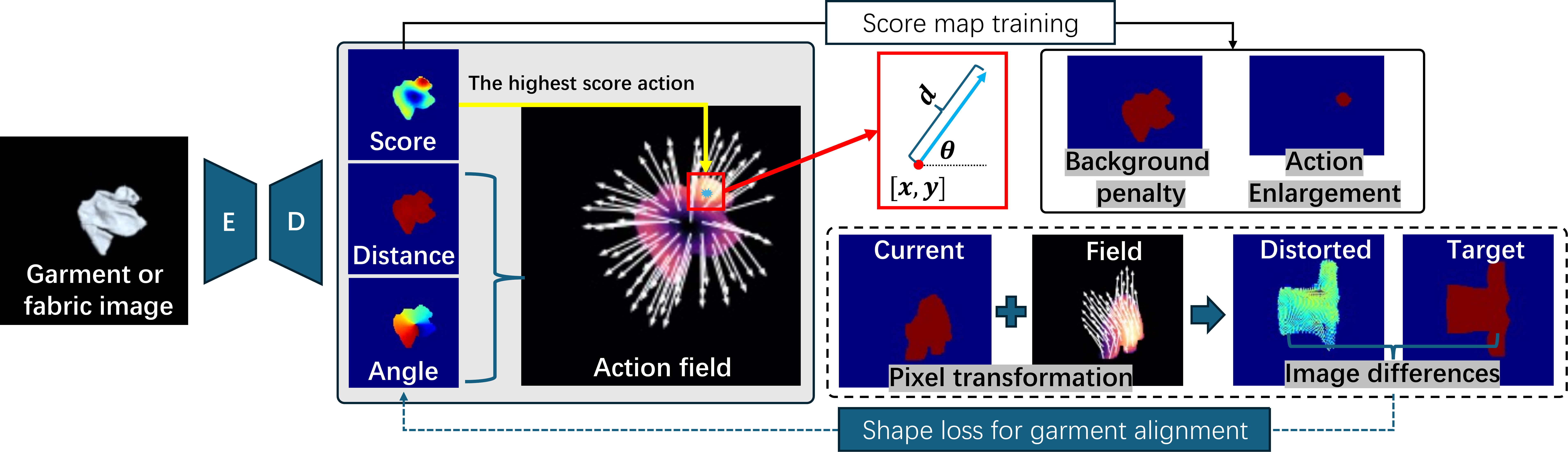}
 \caption{Framework of the proposed dense action generator.}
 \label{fig:pipeline}
\end{figure*}

Based on the action definition in Section~\ref{sec:action-representations}, we design a dense action generator for garment manipulation, as depicted in Fig.~\ref{fig:pipeline}. The model's input is the captured RGB image of the garment or fabric, which is processed by an encoder-decoder to extract features. The output head then produces a 3-channel image that represents the action of the manipulator. Each pixel in the output corresponds to a robot action, with the three channels indicating the action's score, direction, and distance. Using this 3-channel image, we can construct a pixel-wise action field as shown in Fig.~\ref{fig:pipeline}. 

\subsubsection{Model backbone}
The input of the model consists of an RGB image. The features are then extracted using a ResNet-18 \cite{he2016deep} backbone, which functions as the encoder $\mathbf{E}$. The feature maps from different stages of ResNet-18 are then up-scaled and progressively concatenated to match the resolution of the input images. This process forms the decoder $\mathbf{D}$, which provides feature maps enriched with both semantic information and fine-grained details, essential for the output head. The $\mathbf{E}$ and $\mathbf{D}$ constitute the model backbone, as shown in Fig.~\ref{fig:pipeline}. Given that the input images are restricted to garment categories, both a ResNet-based backbone and a U-Net framework \cite{chen2023autobag, chen2023learning} exhibit adequate generalization. We adopt the ResNet as the backbone primarily for its high-capacity feature extraction enabled by skip connections. This design helps mitigate overfitting by residual mappings, which facilitates gradient flow and stable optimization. In this setting, ResNet mainly contributes to faster convergence and reduced training time. 

\subsubsection{Output heads}
The output head is attached to the last-stage feature maps of the model backbone to generate a 3-channel image. Each distinct channel corresponds to a different aspect of the action. The score map indicates the possibility of a pixel location being the start point of the action, the angle and distance map specify the moving direction and magnitude of the action, respectively. By combining the pixel values from these three channels at the same location, a manipulation action can be determined. 

\paragraph{Score head}\label{sec:scorehead} 
The score head is constructed by three stages of cascaded layers: the fully convolutional layer, the ReLU layer, and the Squeeze-and-Excitation block (SEBlock) \cite{hu2018squeeze}. While the former two layers are used to integrate features effectively, SEBlock is introduced to enhance the score map head by recalibrating channel-wise responses. This helps the model focus on more relevant features and improves feature discrimination without the need for additional regularization layers. As a result, the network achieves precise localization of high-score regions that are suitable for smoothing or aligning the fabric.

The architecture of each stage in the score map head is noted as:
\[
\text{Conv2d}(\text{C}_{i-1}, \text{C}_i, \text{$kernel=3$})
\;\rightarrow\;
\text{ReLU}
\;\rightarrow\;
\text{SEBlock}(\text{C}_i),
\]
where $C_{\times}$ represents the channel number of the feature maps, and the kernel size of the 2D convolutional layer is $kernel=3$. Here, $i$ denotes the stage index in the score map head.

\paragraph{Distance head}\label{sec:lengthhead} 
The moving distance of the end-effector $d$ is calculated by multiplying the base unit $d_b$ (in pixels) with a scale factor $s$:
\begin{equation}
    d = s \times d_b.
\end{equation}

\noindent The scale factor $s$ is defined as:
\begin{equation}
    s = d_{\text{scale}} \cdot \text{sigmoid}(x) + d_{\text{offset}},
\end{equation}
where $x$ is the output of the distance head. The $d_{\text{scale}}$ and $d_{\text{offset}}$ are used to linearly scale and shift the sigmoid function. This formulation constrains $s$ within a certain range, ensuring that the calculated distance remains within the robot's physical limitations and operational workspace, as long as $d_b$, $d_{\text{scale}}$, and $d_{\text{offset}}$ are selected appropriately. In the experiments presented in Section~IV, $d_b = 10$, $d_{\text{scale}} = 2.75$, and $d_{\text{offset}} = 0.25$ are chosen.

\paragraph{Angle head}\label{sec:anglehead}
For the angle map, we split the predicted elements into $\sin \theta$ and $\cos \theta$, to avoid ambiguity of the angle periodicity. The angle prediction head consists of two hidden layers followed by two separate heads with tanh activations. The details of the angle head are as follows:
\[
\begin{array}{c}
    \left.
    \begin{array}{c}
        \text{Conv2d}(\text{$C_{in}$}, \text{$C_{out}$}, \text{$kernel=1$}) \rightarrow \text{ReLU} \\
        \downarrow
    \end{array}
    \right\} 
    \times 2 
    \\
    
    \begin{array}{c}
        \text{\text{$SIN({\theta})$}: Conv2d}(\text{$C_{in}$}, \text{$C_{out}=1$}, \text{$kernel=1$}) \rightarrow \text{Tanh}, \\
        \text{\text{$COS({\theta})$}: Conv2d}(\text{$C_{in}$}, \text{$C_{out}=1$}, \text{$kernel=1$}) \rightarrow \text{Tanh}.
    \end{array}
\end{array}
\]

\subsection{Loss Design}\label{sec:lossdesign}
The proposed loss function consists of two components: the score loss, which enables the network to predict a score map for selecting pulling start points, and the angle and distance loss, which guides the network to output pulling directions and distances. Additionally, the shape loss is added as an auxiliary supervision for the alignment task. 

\subsubsection{Score loss}\label{sec:scoreloss}
In previous methods \cite{zeng2018learning, ha2022flingbot, canberk2023cloth}, they only use a single point score and mask out the others in the regression loss calculation. These methods can easily result in potential overfitting during training, leading to the focus drifting from the foreground area and unintentionally highlighting the background. Consequently, it is important to incorporate additional pixel information, specifically the neighboring pixels surrounding the labeled point within the garment region, into the loss calculation. 
\begin{figure}[tbp]
    \vspace{0.20em}
    \centering
    \begin{subfigure}{0.22\columnwidth}
        \includegraphics[width=\textwidth]{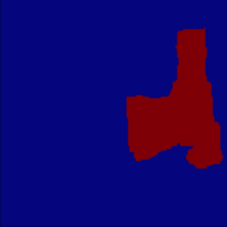}
        \caption{}
        \label{fig:enlargedactioncloth}
    \end{subfigure}
    \begin{subfigure}{0.22\columnwidth}
        \includegraphics[width=\textwidth]{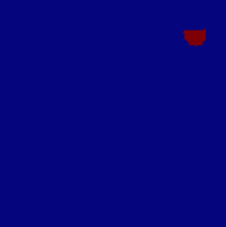}
        \caption{}
        \label{fig:enlargedactionaction}
    \end{subfigure}
    \begin{subfigure}{0.22\columnwidth}
        \includegraphics[width=\textwidth]{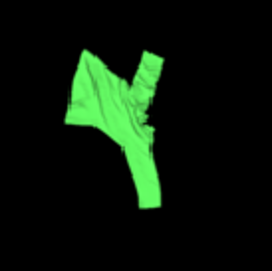}
        \caption{}
        \label{fig:failedstateinput}
    \end{subfigure}
    \begin{subfigure}{0.22\columnwidth}
        \includegraphics[width=\textwidth]{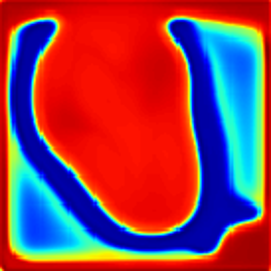}
        \caption{}
        \label{fig:failedscoremap}
    \end{subfigure}
    \caption{(a) shows the garment mask, and (b) shows the enlarged action areas that are filtered by the garment mask. (c) is the input image and (d) is the generated score map without action enlargement, which incorrectly focuses on the background and can not differentiate the areas in the garment or fabric.}
\end{figure}

Unlike the previous score loss design \cite{zeng2018learning, ha2022flingbot, canberk2023cloth}, which only considers a single score value during the sampling, our score loss design incorporates the prior knowledge about robot manipulation of a fabric:
\begin{enumerate}[label=(\arabic*)]
  \item The movement results of fabric manipulation should have consistency within a small local area. 
  \item The pulling action should preferably originate within the boundary of the fabric. 
\end{enumerate}
These conditions reflect the reality of the fabric manipulation task and also help to improve the score regression performance, particularly in addressing the extreme imbalance issue between the foreground and background pixel numbers. Notably, each training sample includes a score label for only a single pixel in the simulated image.

According to constraint (1), adjacent points on the fabric generally lead to similar manipulation results due to the local homogeneity and continuity of the deformable material. To satisfy constraint (2), the pulling start points should be accurately confined within the garment by applying a segmentation mask (Fig.~\ref{fig:enlargedactioncloth}). Based on these considerations, we assign the same score value to the neighboring pixels around the sampled point, as long as they lie within the garment region (Fig.~\ref{fig:enlargedactionaction}). The score is computed based on the difference in the garment's state before and after applying the corresponding action. 

We effectively increase the area of the foreground by convolving the mask of the score labeled action area $M_{Action}$ with a predefined circular kernel $K$, and then applying binary thresholds to obtain the enlarged action mask $M_{EAction}$:
\begin{equation}
    \begin{aligned}
    K_{i, j} &= \begin{cases} 
        1 & \text{if } i^2 + j^2 \leq r^2, \text{ for } i, j \in [-r, r] \\
        0 & \text{otherwise}
    \end{cases}, \\
    M_{EAction} &= \begin{cases}
        1 & \text{if } (M_{Action} \ast K) > 0 \\
        0 & \text{otherwise}
    \end{cases},
    \end{aligned}
\end{equation}
where the 2D circular kernel $K$ is with size $(2r+1) \times (2r+1)$, and $r = 3$, and $M_{Action} \ast K$ denotes the convolution of the action mask $M_{Action}$ with the kernel $K$. One sample of a $M_{EAaction}$ is shown in Fig.~\ref{fig:enlargedactionaction}.

Then, the adjusted loss for action, $L_{a}$, is calculated as: 
\begin{align}
L_{a} &= \text{Mean}\Big(\text{SmoothL1Loss}\Big(\tilde{I}_{s} \odot M_{EAction}, 
    \nonumber \\
     &\quad gt_{s} \cdot M_{EAction}\Big)\Big),
\end{align}
where $gt_s$ is a single ground truth score value from the training data, and $\tilde{I}_{S}$ is the predicted action score map. 

Meanwhile, the score map (Fig.~\ref{fig:failedscoremap}) generated for the input (Fig.~\ref{fig:failedstateinput}) by the single pixel regression erroneously focuses on the background, failing to accurately and clearly segment the garment. Based on this pre-training result and given the constraint (2), the minus scores $gt_{b}=-1$ are assigned to the background pixels $M_{Background}$, which also guides the training process by penalizing the background areas. The background loss, $L_b$, is calculated as:
\begin{align}
L_{b} &= \text{Mean}\Big(\text{SmoothL1Loss}\Big(\tilde{I}_{s} \odot M_{Background},  
    \nonumber \\
     &\quad gt_{b} \cdot M_{Background} \Big)\Big).
\end{align}

Instead of constraining only a single action point \cite{zeng2018learning, ha2022flingbot, canberk2023cloth}, our final score loss includes the regression loss of each pixel from both the enlarged action and background area. Specifically, the score loss $L_{score}$ is the weighted sum of $L_a$ and $L_b$, whose contribution is adjusted by $\lambda_b=0.001$. The $L_{score}$ is calculated as:
\begin{equation}
L_{score} = L_{a} + \lambda_b \cdot L_{b}.
\end{equation}
Through these processes, the foreground area is effectively enlarged and the background information is also considered, which helps mitigate overfitting caused by the class imbalance. 

\subsubsection{Angle and distance loss}\label{sec:angleandlengthloss}
Both the angle loss $L_{\text{angle}}$ and the distance loss $L_{\text{distance}}$ are computed by averaging the regression errors between the predicted and ground truth values in the angle map $I_A$ and the distance map $I_D$. These losses are calculated only at the pixel locations specified by the action mask $M_{\text{EAction}}$.

The angle loss, $L_{angle}$, based on the angle prediction in Section~\ref{sec:anglehead}, requires the regression of two corresponding values, $\sin(\theta)$ and $\cos(\theta)$, and a penalty part to regularize them to satisfy the unit circle constraint:
\begin{align}
    L_{sin} &=  \text{Mean}(\text{SmoothL1Loss}(\tilde{I}_{sin} \odot M_{EAction}, \nonumber \\
    &\quad gt_{sin} \cdot M_{EAction})), \\
    L_{cos} &=  \text{Mean}(\text{SmoothL1Loss}(\tilde{I}_{cos} \odot M_{EAction}, \nonumber \\
    &\quad gt_{cos} \cdot M_{EAction})), \\
    L_p &= \text{Mean}((\tilde{I}_{sin})^2 + (\tilde{I}_{cos})^2 - 1)^2 \odot M_{EAction}), \\
    L_{angle} &= L_{sin} + L_{cos} + \lambda_p L_p ,
\end{align}
where \(L_{sin}\) and \(L_{cos}\) are the regression losses for the predicted angle images, $\tilde{I}_{sin}$ and $\tilde{I}_{cos}$, respectively, under the supervision of the triangle values, $gt_{sin}$ and $gt_{cos}$ of the ground truth angle. Finally, the angle regression loss $L_{angle}$ is their weighted sum with the unit norm penalty $L_p=1.0$ with weight $\lambda_p$. 

The calculation of the distance loss, $L_{distance}$, is also the regression loss between the predicted distance map, $\tilde{I_d}$, and the labeled ground truth distance values, \(gt_d\):
\begin{align}
    L_{distance} &= \text{Mean}\Big(\text{SmoothL1Loss}\Big(\tilde{I_d} \odot M_{EAction}, 
    \nonumber \\
    & gt_d \cdot M_{EAction}\Big)\Big),
\end{align}

\subsubsection{Shape loss}\label{sec:shapeloss}
\begin{figure}[tbp]
    \vspace{0.20em}
    \centering
    \begin{subfigure}{0.24\columnwidth}
        \includegraphics[width=\textwidth]{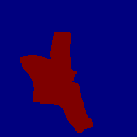}
        \caption{Current state}
        \label{fig:shape-loss-current-state}
    \end{subfigure}
    \begin{subfigure}{0.24\columnwidth}
        \includegraphics[width=\textwidth]{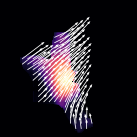}
        \caption{Action field}
        \label{fig:shape-loss-action-field}
    \end{subfigure}
    \begin{subfigure}{0.24\columnwidth}
        \includegraphics[width=\textwidth]{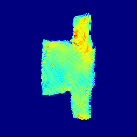}
        \caption{After state}
        \label{fig:shape-loss-after-state}
    \end{subfigure}
    \begin{subfigure}{0.24\columnwidth}
        \includegraphics[width=\textwidth]{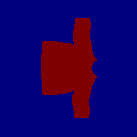}
        \caption{Target state}
        \label{fig:shape-loss-target-state}
    \end{subfigure}
    \caption{The state images and the action field involved in calculating the shape loss. (a) shows the mask of the garment in its current state. (b) is the action field determined by the predicted angle and distance map. (c) is the result of applying the dense actions (b) to each pixel in the mask of (a). (d) is the mask representing the target state of the garment in the alignment task. }
    \label{fig:shape-loss}
\end{figure}

In addition to supervising the training with angles and distances labeled by humans or simulators, our model is also supervised using the desired final shape of the garment for the alignment task. The calculation components of the proposed shape loss are shown in Fig.~\ref{fig:shape-loss}. 

In the garment alignment task, the final goal is deterministic and can be represented by a garment mask in a predefined, fully flattened pose. Based on the predetermined target state, we propose a novel loss function that compares the discrepancy between the garment mask of the target state $I_{target}$ (Fig.~\ref{fig:shape-loss-target-state}) and that of the approximate state $I_{distort}$ (Fig.~\ref{fig:shape-loss-after-state}), where $I_{distort}$ is generated through the pixel level transformation $\mathcal{D}$ based on the densely generated actions. 

\begin{algorithm}[tbp]
\caption{Distorted Garment Image Generation}
\label{alg:forward_warp}
\begin{algorithmic}[1]

\Require Current state mask $I_{current}$, predicted angle map $\tilde{I}_{\sin}$ and $\tilde{I}_{\cos}$, and predicted distance map $\tilde{I}_d$ (all in $\mathbb{R}^{H \times W}$)
\Ensure Distorted mask $I_{distort} \in \mathbb{R}^{H \times W}$

\AlgPhase{1. Compute transformations based on the action field}
\State $P_{{src}} = (x, y)$ denotes the grid coordinates of $I_{current}$

\State $\Delta x \leftarrow \tilde{I}_d \odot \tilde{I}_{\cos}$
\State $\Delta y \leftarrow \tilde{I}_d \odot \tilde{I}_{\sin}$

\State $P_{{target}} \leftarrow P_{{src}} + (\Delta x, \Delta y)$

\AlgPhase{2. Compute value weights for neighboring pixels}
\State $P_0 \leftarrow \lfloor P_{{target}} \rfloor$ 

\State $P_1 \leftarrow P_0 + 1$
\State Let $P_{{target}} = (x_t, y_t)$, $P_0 = (x_0, y_0)$, $P_1 = (x_1, y_1)$
\State $\omega_{00} \leftarrow (x_1 - x_t) \odot (y_1 - y_t)$
\State $\omega_{10} \leftarrow (x_t - x_0) \odot (y_1 - y_t)$
\State $\omega_{01} \leftarrow (x_1 - x_t) \odot (y_t - y_0)$
\State $\omega_{11} \leftarrow (x_t - x_0) \odot (y_t - y_0)$

\AlgPhase{3. Compute target values by the weighted sum of the neighboring pixels}
\State $I_{{distort}} \leftarrow \mathbf{0}$
\For{each offset $(i, j) \in \{(0,0), (1,0), (0,1), (1,1)\}$}
    \State $P_{ij} \leftarrow P_0 + (i, j)$
    \State $I_{{distort}}[P_{ij}] \mathrel{+}= I_{{current}} \odot \omega_{ij}$
\EndFor

\end{algorithmic}
\end{algorithm}

The transformation function $\mathcal{D}$, which maps the current mask $I_{current}$ (Fig.~\ref{fig:shape-loss-current-state}) to the approximate state $I_{distort}$, must be differentiable to allow gradients to backpropagate through the network. To achieve this, we implement $\mathcal{D}$ as a differentiable forward warping process, which is detailed in Algorithm \ref{alg:forward_warp}. Specifically, for each pixel in $I_{current}$, we first compute its floating-point target coordinate based on the action field that is constructed by [$\tilde{I}_{sin}$, $\tilde{I}_{cos}$, $\tilde{I}_{d}$]. Then, the pixel's value is distributed to the grid neighborhood surrounding its target coordinate, with weights determined bilinearly. The final intensity at each location in $I_{distort}$ is the sum of all contributions scattered to it. 

The predicted angle and distance maps can be combined into a single action field image, as shown in Fig.~\ref{fig:shape-loss-action-field}. In both simulated and real-world physics, each individual action depicted in Fig.~\ref{fig:shape-loss-action-field} cannot independently transform a single point due to the interconnected forces of adjacent points. However, the aggregate effect of these actions can approximate the overall manipulation results if all points are moved according to all predicted actions. In other words, the overall transformations based on the action field provide valuable insight into the efficacy of the predicted actions, and these transformations can be observed as a displacement field $\mathcal{D}$. $\mathcal{D}$ denotes the horizontal and vertical pixel movement offsets $[\Delta x, \Delta y]$ to formulate a new image. To this end, we propose the shape loss during the training: 
\begin{align}
    L_{shape} &= MSE(I_{target}, I_{distort}),\\
    I_{distort} &= \mathcal{D}(I_{current}).
\end{align}
The shape loss $L_{shape}$ is a classification loss and computed using Mean Squared Error (MSE) to assess the discrepancy between the target mask $I_{target}$ and the distorted image $I_{distort}$. 

\subsubsection{Total loss}
The total loss is the sum of the action score loss and the action parameters regression losses (angle and distance). There is an additional shape loss for alignment tasks. Detailed formulation of the total loss is illustrated in Algorithm \ref{alg:main_opt_academic} and can be expressed as:
\begin{align}
    L_{total} = L_{score} + \lambda_a L_{angle} + \lambda_d L_{distance} + \lambda_s L_{shape},
    \label{eq:total_loss}
\end{align}
where both $\lambda_{a}$ and $\lambda_{d}$ are set to 0.1 to balance their contributions. $\lambda_s$ is set to zero during the smoothing task, and to 25.0 for the alignment task to minimize excessive focus on the target shape, which could lead to overfitting.

\begin{algorithm}[tbp]
\caption{Total Loss Calculation with Multiple Components}
\label{alg:main_opt_academic}
\begin{algorithmic}[1]
\Require Model $M_\theta$, training Set $\mathcal{T}_{train}$, and loss weights $\{\lambda_b, \dots, \lambda_s\}$
\Ensure Total Loss $L_{{total}}$
    \State $\{I_{{RGB}}, I_{{current}}, I_{{target}}, M_{{Action}}, gt_s, gt_{\sin}, \dots\} \sim \mathcal{T}_{{train}}$
    \State $\tilde{I}_s, \tilde{I}_{\sin}, \tilde{I}_{\cos}, \tilde{I}_d \leftarrow M_\theta(I_{{RGB}})$
    
    \State $M_{{EAction}} \leftarrow \text{Binarize}((M_{{Action}} \ast K) > 0)$
    \State $M_{{Background}} \leftarrow \bm 1 - I_{{current}}$
    \State $gt_b \leftarrow -1 $
    
    \State $L_{{score}} \leftarrow \text{ScoreLoss}(\tilde{I}_s, gt_s, gt_b, M_{{EAction}}, M_{{Background}})$
    \State $L_{{angle}} \leftarrow \text{AngleLoss}(\tilde{I}_{\sin}, \tilde{I}_{\cos}, gt_{\sin}, gt_{\cos}, M_{{EAction}})$
    \State $L_{{distance}} \leftarrow \text{DistanceLoss}(\tilde{I}_d, gt_d, M_{{EAction}})$
    
    \State $I_{{distort}} \leftarrow \mathcal{D}(I_{{current}}, \{\tilde{I}_{\sin}, \tilde{I}_{\cos}, \tilde{I}_d\})$
    \State $L_{{shape}} \leftarrow \text{ShapeLoss}(I_{{distort}}, I_{{target}})$

    \State $L_{{total}} \leftarrow L_{{score}} + \lambda_a L_{{angle}} + \lambda_d L_{{distance}} + \lambda_s L_{{shape}}$
    
\end{algorithmic}
\end{algorithm}

Our training loss aggregates multiple supervision signals that directly supervise fully determined action parameters, i.e., starting location, direction, and distance, rather than the manipulation scores solely. Optimizing this objective enables the model to output actions closer to the ground truths or actions that can reduce the IoU values. We determine the weight balance factors, $[\lambda_b, \lambda_p, \lambda_a, \lambda_d]$, by grid search. In a preliminary study using 1024 samples held out from the training set, each value was iterated over $\{0.001,0.01,0.1,1.0,10.0 \}$. For efficiency considerations, we evaluated 200 stratified random combinations instead of the full point grid, i.e., totaling $5^4=625$ points. The final weight values were chosen based on the best performance observed during these experiments.

\subsection{Data Collection}\label{sec:datacollection}
Considering the high-dimensional nature of the fabric state and the action space, training the model presents significant challenges, requiring a large volume of high-quality data. Thus, we collected the training data in the simulator to avoid tedious human labeling. 

\subsubsection{Action labeling}\label{sec:actionlabeling}
\begin{figure}[tb]
    \vspace{-0.15em}
    \centering
    \begin{subfigure}{0.22\columnwidth}
        \includegraphics[width=\textwidth]{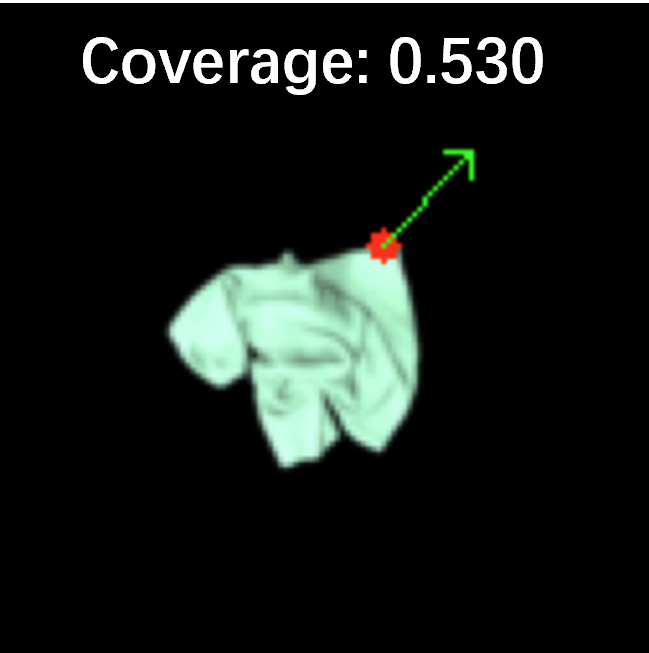}
        \caption{Init state}
        \label{fig:samle-init}
    \end{subfigure} 
    \begin{subfigure}{0.22\columnwidth}
        \includegraphics[width=\textwidth]{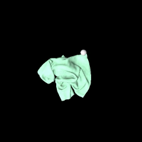}
        \caption{Action start}
        \label{fig:sample-start}
    \end{subfigure}
    \begin{subfigure}{0.22\columnwidth}
        \includegraphics[width=\textwidth]{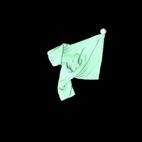}
        \caption{Action end}
        \label{fig:sample-end}
    \end{subfigure}
    \begin{subfigure}{0.22\columnwidth}
        \includegraphics[width=\textwidth]{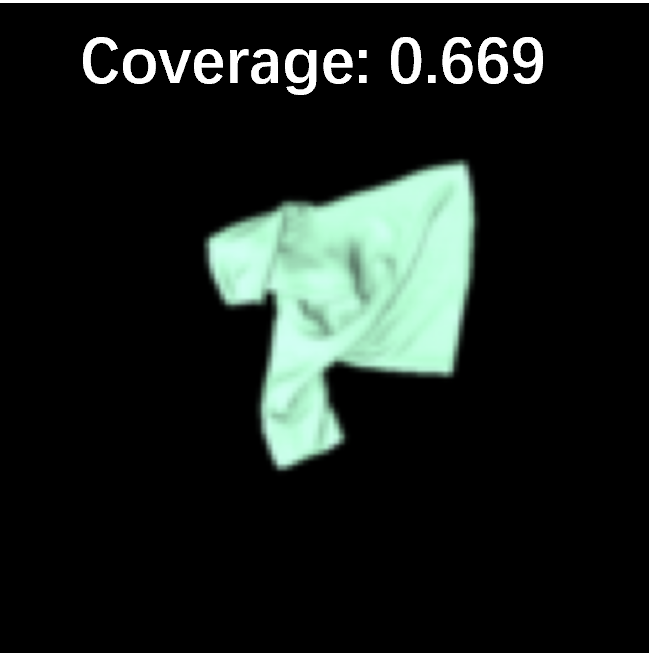}
        \caption{Final state}
        \label{fig:sample-final}
    \end{subfigure}
    \caption{The sampled action and the state changes of a garment during the action labeling. (a) shows the current state of the garment and the randomly generated action. (b) and (c) depict the start and end state snapshots of attaching the ``pull'' action. (d) presents the final state.}
    \label{fig:sample}
\end{figure}

Training samples are labeled using simulators described in \cite{ha2022flingbot} and \cite{canberk2023cloth}. These simulators are built on the SoftGym \cite{lin2021softgym} environment that uses the NVIDIA FleX wrapper PyFlex \cite{li2018learning}. The randomly sampled action is defined in Section~\ref{sec:action-representations} and the state changes of a garment during the action labeling are shown in Fig.~\ref{fig:sample}. The labeling process involves recording pre- and post-action garment states, with state parameter change as the action score. Specifically, smoothing is quantified by the coverage index (ratio of current to fully-flattened area), while alignment is measured by the alignment index (Intersection over Union (IoU) between current and target garment masks). Note that this target state is predefined according to the category of the garments, e.g., a fully unfolded long sleeve with its neck pointing to the right. The objective of our alignment task is then to move the garment towards this predefined target configuration. These coverage and alignment indexes also serve as the evaluation metrics for the experiments, with higher values indicating better performance.

\begin{figure}[tb]
    \centering 
    \begin{subfigure}{0.44\columnwidth}
        \includegraphics[width=\textwidth]{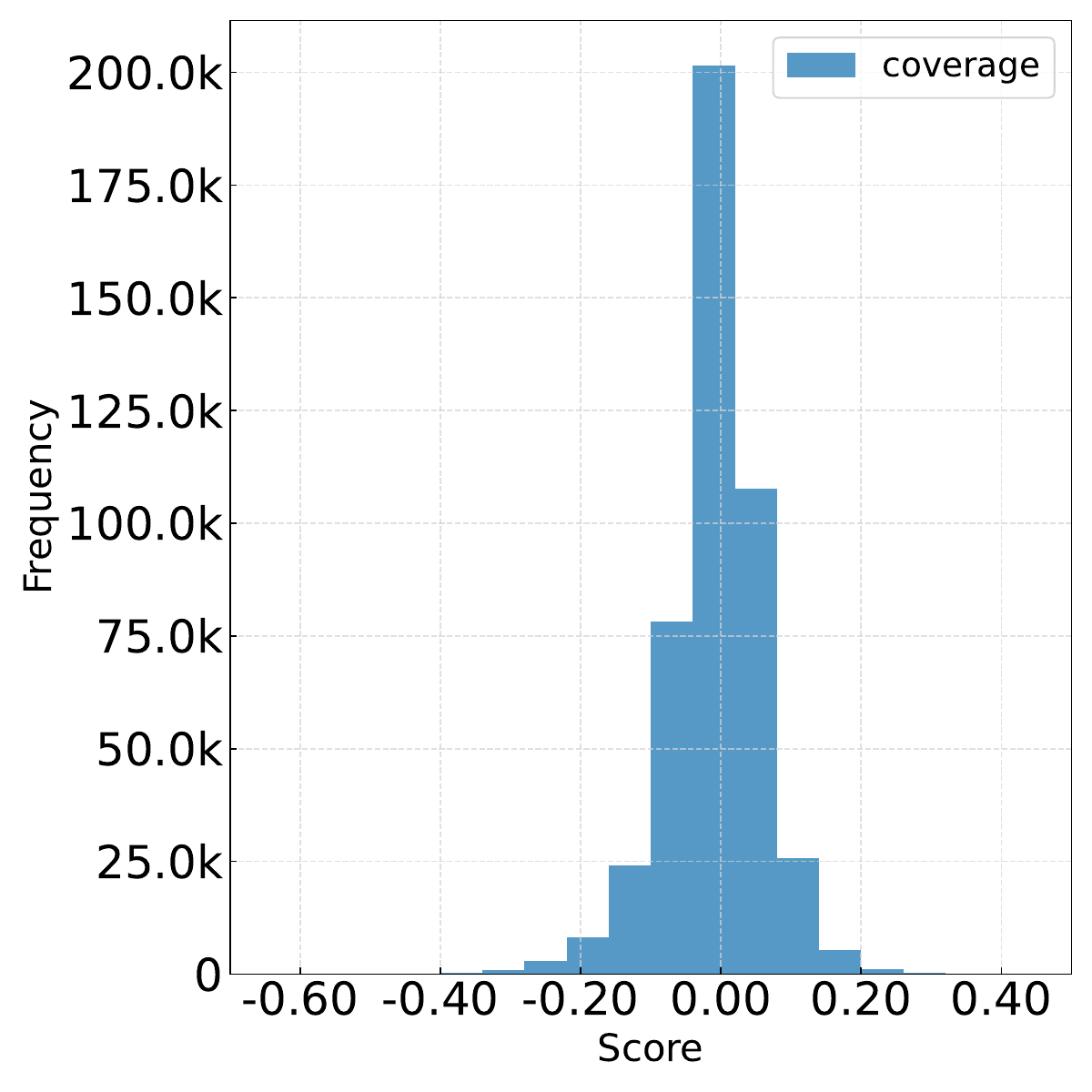}
        \caption{Initial samples}
        \label{fig:refine-before}
    \end{subfigure} 
    \begin{subfigure}{0.44\columnwidth}
        \includegraphics[width=\textwidth]{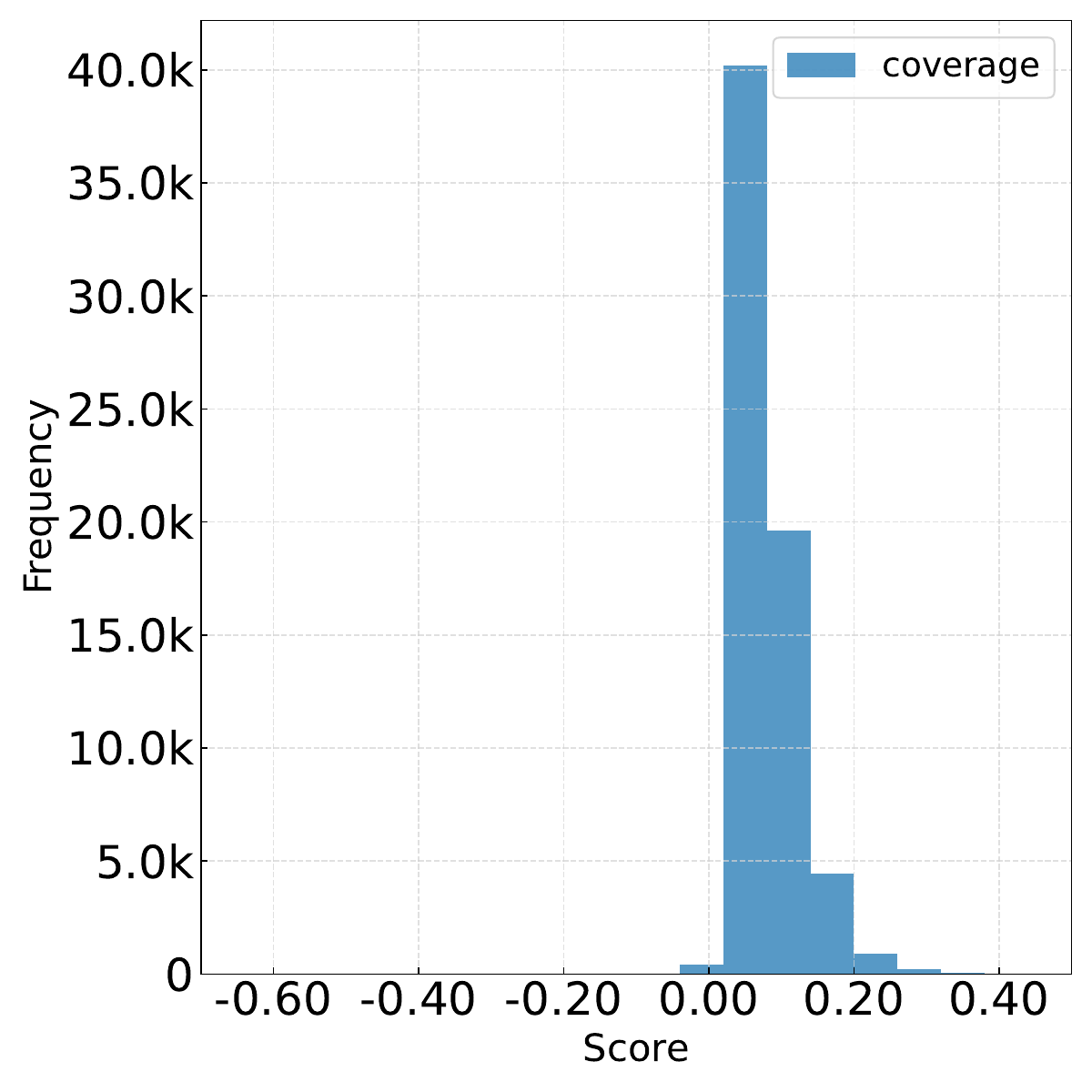}
        \caption{Refined samples}
        \label{fig:refine-after}
    \end{subfigure} \\
    \caption{Distributions of the reward score (coverage index) before (a) and after (b) the refining process.}
    \label{fig:refine}
\end{figure}

To generate actions densely, our model outputs the score, angle, and distance maps in parallel. Samples with low reward scores should be refined to prevent adverse effects during training. To achieve this, the sampled actions are pruned based on a predefined task-specific metric threshold, e.g., changes in coverage index. Samples with negative metrics are discarded, and those with lower values are downsampled, significantly reducing the total sample numbers. Changes in the distribution of the dataset are illustrated in Fig.~\ref{fig:refine}. 

\subsubsection{Statistic details}
Combining the image with the corresponding action configurations and the recorded scores, we present the garment manipulation dataset specially for the ``pull'' action. Statistically, we generate 65,000 training samples with the ``long sleeves'' clothing category for the garment smoothing tasks and 27,000 training samples for the alignment tasks. For evaluation purposes, 400 newly generated scenarios per clothing category (long sleeves, pants, skirts, dresses, and jumpsuits) are created, each containing garments in crumpled states. 

\section{Experiments}
In this section, we first compare our model with the state-of-the-art methods using simulated environments. Results of real-world experiments are also provided to demonstrate the effectiveness of the proposed method. These results help evaluate the domain gap between the simulated images and the real-world image of the garment or fabric. 

\subsection{Dataset Experiments}\label{sec:datasetexperiments}
We compare our dense action generator with the two-stage method VCD \cite{lin2022learning} and the single-stage methods, i.e., FlingBot \cite{ha2022flingbot} and Cloth Funnels \cite{canberk2023cloth}. The VCD is a milestone work of the two-stage garment manipulation methods, which models the garment as a particle-based graph and predicts its realistic movements under external force. The FlingBot is the pioneering work adopting the spatial action map strategy for garment smoothing, featuring an easier structure and higher efficiency compared to the Cloth Funnels. The Cloth Funnels adopted factorized objective functions.

\subsubsection{Quantitative results}\label{sec:quantityresults}

\begin{table}[tbp]
    \centering
    \caption{Evaluation Results on the Smoothing Task}
    \begin{tabular}{c ccc c}
        \toprule
        \multirow{2}{*}{Methods} & \multicolumn{3}{c}{Coverage} & \multirow{2}{*}{Time (ms)} \\
        \cmidrule(lr){2-4}
                                 & Hard & Easy & Rectangular & \\ 
        \midrule
        VCD \cite{lin2022learning}       &--  &--  &0.876  &5004.0  \\
        FlingBot \cite{ha2022flingbot}        &0.788  &0.920  &0.893  &47.3  \\ 
        Cloth Funnels \cite{canberk2023cloth} & 0.792      & 0.932  & 0.921  & 186.0      \\ 
        Ours                                   & 0.854      & 0.989 & 0.944  & 22.6       \\
        \bottomrule
    \end{tabular}
    \label{tab:unfolding}
\end{table}

In Table~\ref{tab:unfolding}, we show the evaluation results of our method and the comparison methods. For each method, we follow the same strategy for collecting the initial crumpled garments, i.e., manipulate the garments with different levels of force, yielding different degrees of crumpling that are categorized as ``Hard" and ``Easy" tasks. The VCD is evaluated by the rectangular fabric that is in a crumpled state. Meanwhile, similar crumpled-state rectangular fabrics are generated for the single-stage methods for comparison. These crumpled-stated garments and cloth are manipulated with the planar action ``pull''. 
Our method achieves a coverage index value of $0.854$ for the ``Hard'' tasks, with a relative increase of $7.83\%$ over the competing method with the best performance, i.e., $0.792$ of the Cloth Funenls, demonstrating its superior performance in wrinkle elimination and garment smoothing. For the ``Easy'' tasks with slightly curmpled garments, our algorithm also achieves the highest coverage index of $0.989$. Compared with the two-stage method VCD, all single-stage methods show better results in the rectangular fabric case, which presents the benefit of generating the action directly. Regarding efficiency, we report the per-step running time measured on a single NVIDIA A6000 GPU. The VCD requires re-running the mesh graph model to predict the effect of several randomly sampled actions, which have a much heavier computational burden than single-stage methods. We select the best action from 50 randomly sampled actions, which takes approximately 5.0 seconds. While conventional spatial action map methods require multiple forward passes of the model for a sequence of transformed inputs, our architecture generates continuous and dense actions in a single pass. Consequently, our approach operates at $22.6$ ms per generation compared to $186.0$ ms for Cloth Funnels and $47.3$ ms for Flingbot, significantly improving both efficiency and coverage performance in garment smoothing. 

\begin{table}[tbp]
    \centering
    \caption{Evaluation Results on Smoothing Task for Different Garment Types}
    \label{tab:unfolddifferentclothes}
    \begin{tabular}{l c c c c}
        \toprule
        Train Set & Test Set & \makecell{Coverage} & \makecell{Training\\Sample (No.)} & \makecell{Scene \\ (No.)} \\
        \midrule
        \multirow{5}{*}{\makecell{Long
        \\sleeves}} 
            & Long sleeves & 0.854 & 65,000               & \makecell{2000 training \\ 400 testing} \\
            & Pants       & 0.855 & \multirow{4}{*}{N/A}   & \multirow{4}{*}{400 testing} \\
            & Skirt       & 0.915 &                      &                      \\
            & Dress       & 0.917 &                      &                      \\
            & Jumpsuit    & 0.857 &                      &                      \\
        \bottomrule
    \end{tabular}
\end{table}

We further assess the generalization of the proposed model by testing it on diverse garment types, using the same model trained exclusively on ``long sleeves'' samples. The results are presented in Table~\ref{tab:unfolddifferentclothes}, demonstrating that the model achieves effective smoothing outcomes and maintains robustness beyond the training distribution. Among these clothing categories, the ``long sleeves'', ``pants'', and ``jumpsuit'' exhibit nearly identical coverage index values, which are relatively lower than those for other categories. This discrepancy is likely due to the complexities involved in manipulating sleeves and legs.
\begin{table}[tbp]
    \caption{Evaluation Results on the Alignment Task}
    \begin{center}
    \begin{tabular}{c c}
        \toprule
        Methods & IoU \\
        \midrule
        Cloth Funnels \cite{canberk2023cloth} & 0.535 \\ 
        \makecell{Ours} & 0.580 \\
        \bottomrule
    \end{tabular}
    \label{tab:alignmenttask}
    \end{center}
\end{table}

In addition to the smoothing task, we also evaluate our methods by the alignment task, where the objective is to position the garment into a specific target pose while smoothing. By incorporating the shape loss, our model achieves the garment alignment task with the same model structure and returns a higher alignment index, i.e., IoU value, than Cloth Funnels. The results are presented in Table~\ref{tab:alignmenttask}.

\subsubsection{Qualitative  results}\label{sec:qualityresults}

\begin{figure}[tbp] 
 \centering
 \includegraphics[width=0.9\columnwidth]{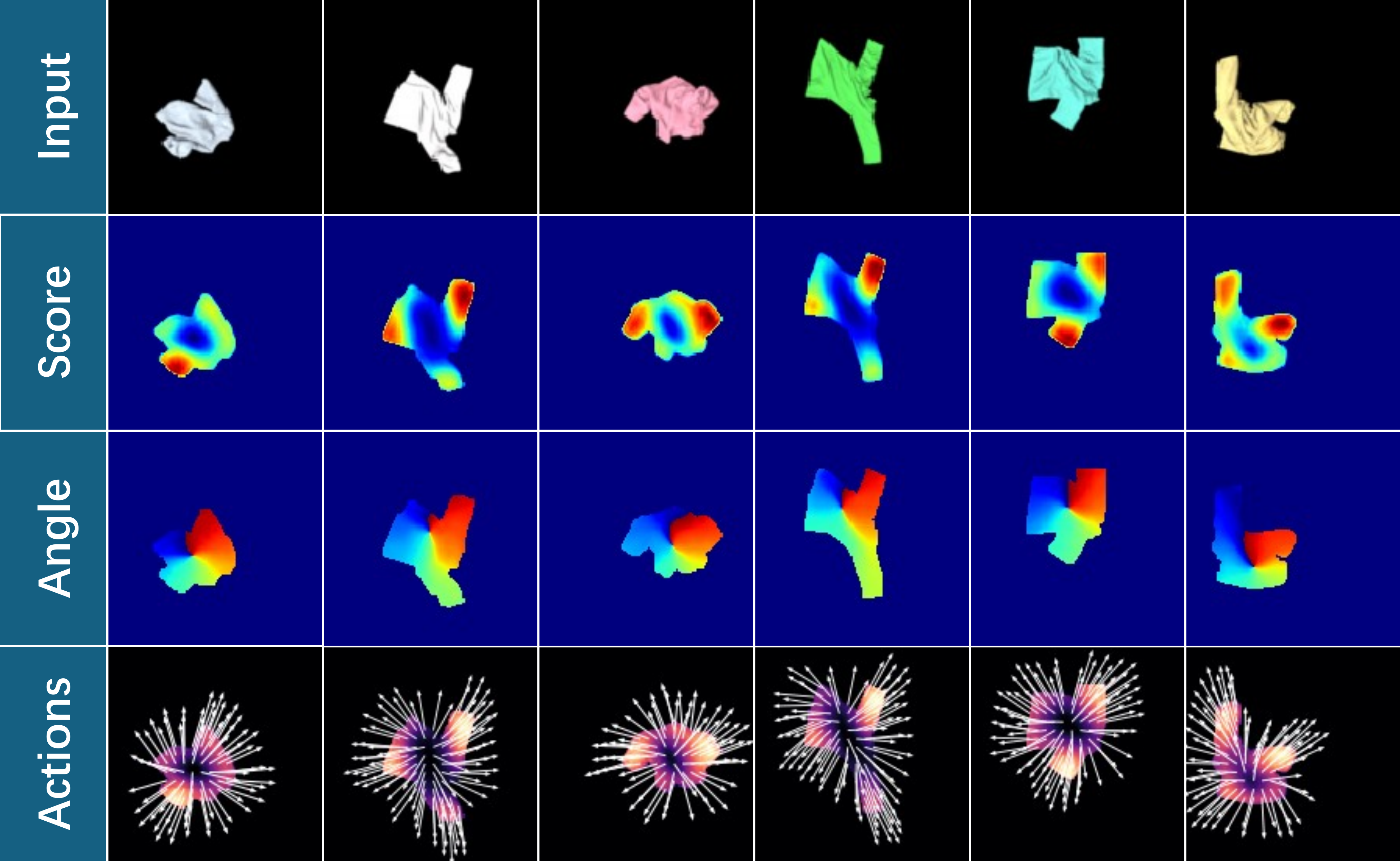}
 \caption{The visualizations of the score map and dense actions, from top to bottom, are the input, the score map, the angle map, and the generated actions (selected at intervals of 7 pixels) within the garment. }
 \label{fig:simulatedcoveragedensevisualization}
\end{figure}

Fig.~\ref{fig:simulatedcoveragedensevisualization} presents the qualitative image results, including score maps, angle maps, and action field visualizations of the model outputs. The score map accurately locates the key region for smoothing, such as the sleeve edge, hem, or neckline, rather than the center areas, where manipulation in any direction tends to generate wrinkles. Meanwhile, the action field consistently indicates a direction from the garment's center towards its mask boundary. This aligns with the radial flattening strategy, where dragging outwards is generally preferred for removing wrinkles. 

\begin{figure*}[htbp] 
 \centering
 \includegraphics[width=0.9\textwidth]{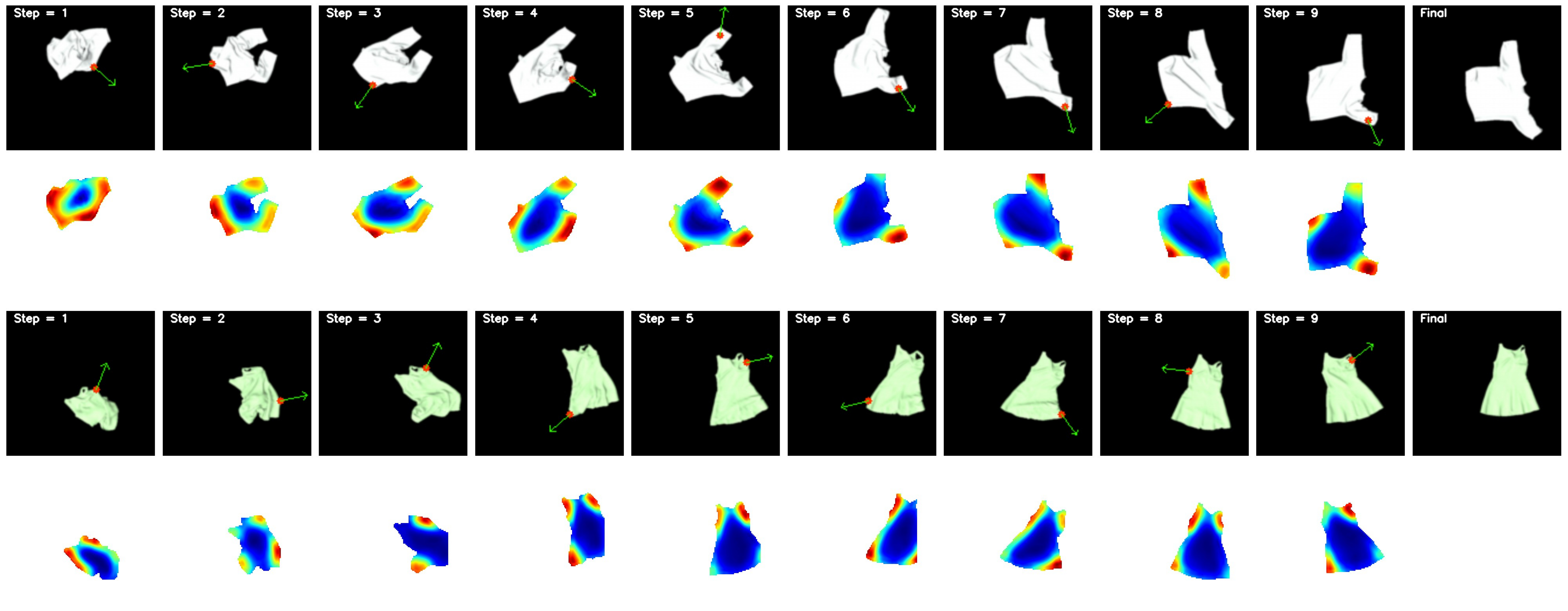}
 \caption{The step-by-step visualizations of the garment smoothing task of the long sleeves and dress.}
 \label{fig:simulatedcoveragesteps}
\end{figure*}

\begin{figure}[htbp]
    \vspace{-2em}
    \centering
    \includegraphics[width=0.9\columnwidth]{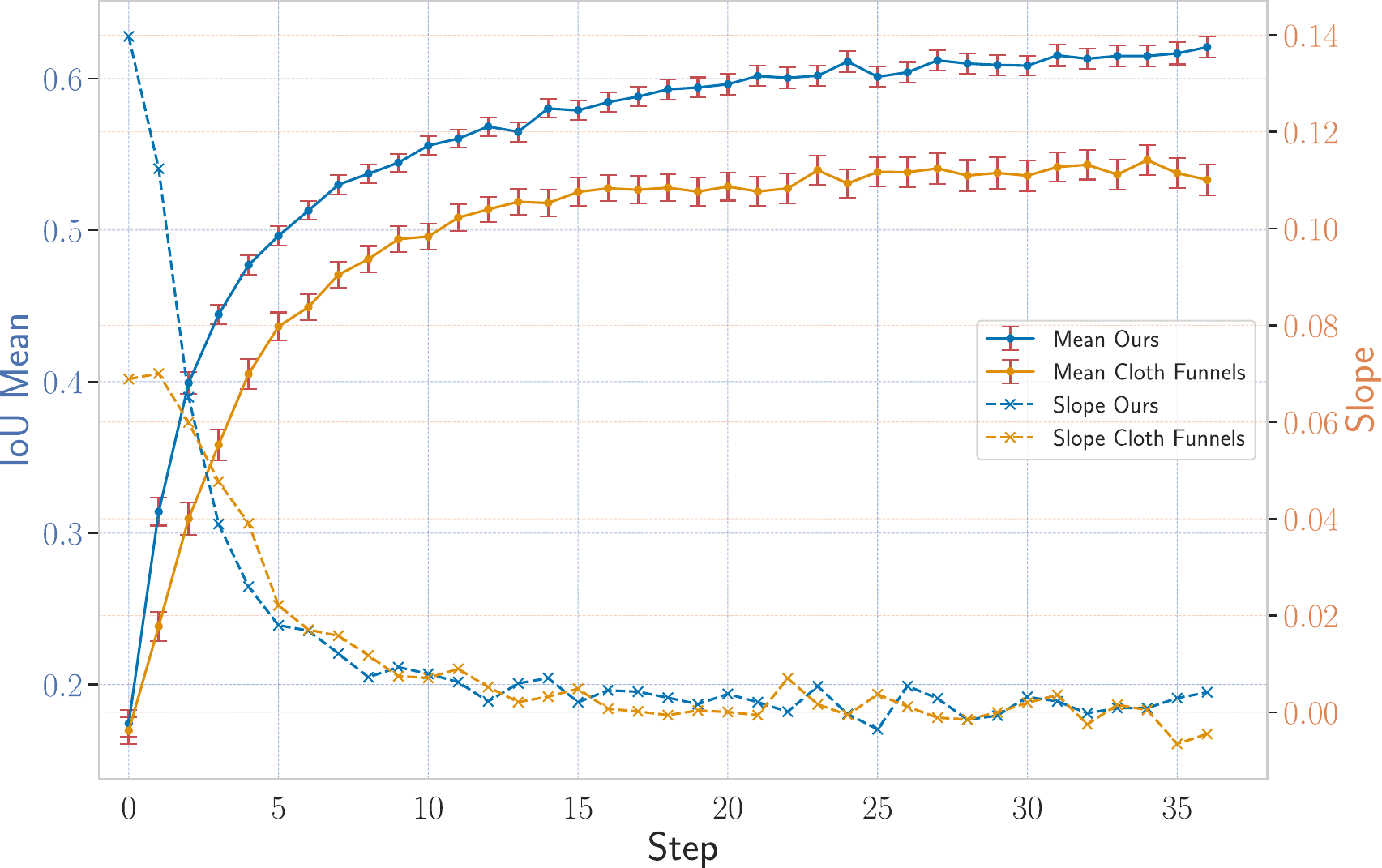}
    \caption{Garment alignment experiment results. The slope indicates the increasing speed of the IoU Mean values.}
    \label{fig:alignment}
\end{figure}

\begin{figure}[htbp] 
\vspace{-0.5em}
 \centering
 \includegraphics[width=0.9\columnwidth]{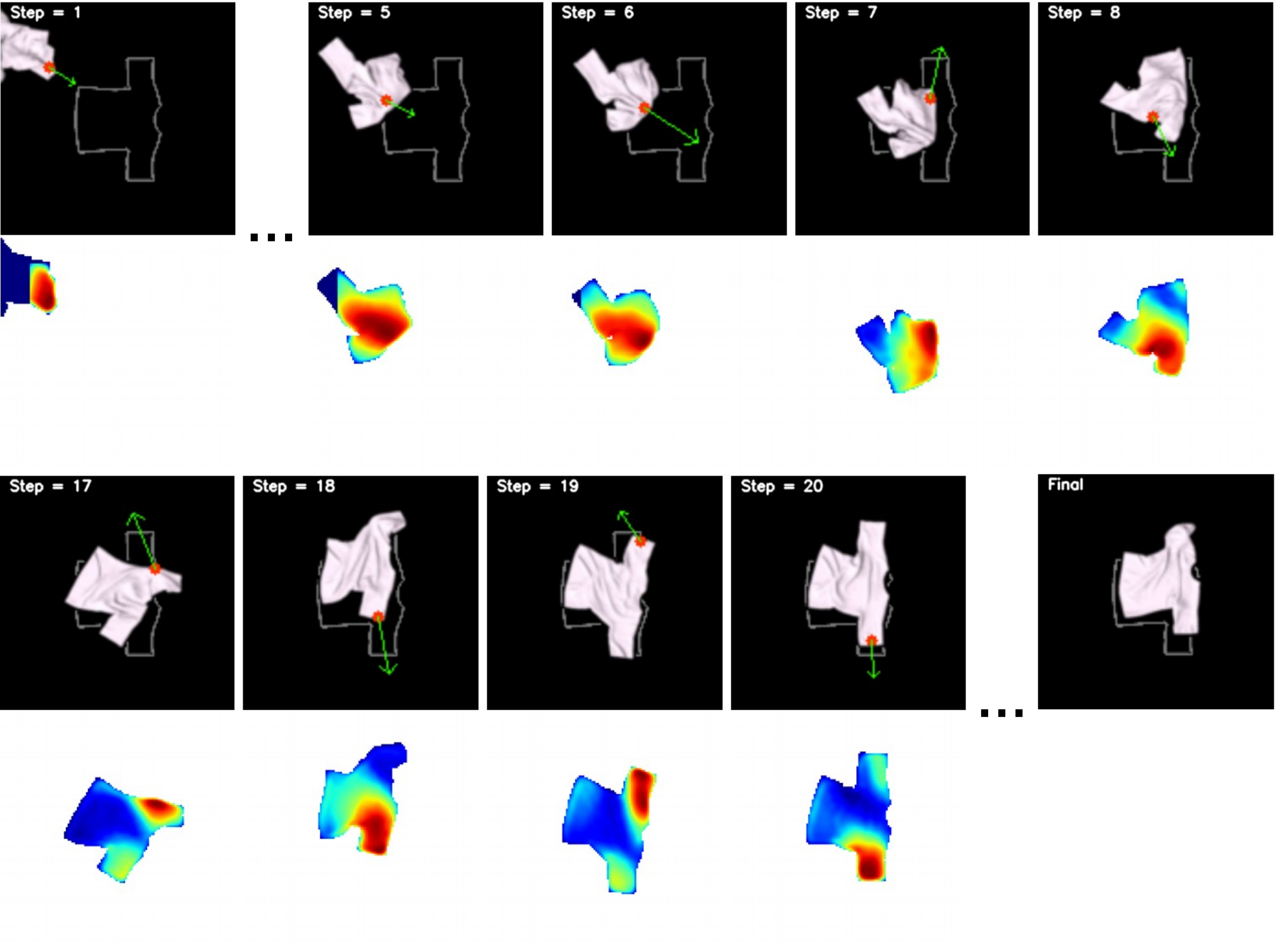}
 \caption{The visualizations of the garment alignment task, with a predefined goal of the neck pointing right. The white line drawings indicate the target state.}
 \label{fig:simulatedalignment}
\end{figure}

Fig.~\ref{fig:simulatedcoveragesteps} shows complete smoothing processes for the long sleeves and dress with actions generated by the same model. Although the model was trained on isolated samples, the smoothing sequences demonstrate its potential for long-horizon manipulation. Initially, the garments are heavily crumpled, and the model smooths the outer edges to increase the overall coverage area. As key regions such as corners, hems, and necklines become exposed, the model progressively shifts its focus to those areas, generating finer actions to eliminate small wrinkles.

For the alignment task, we plot the IoU value after each ``pull'' action, as shown in Fig.~\ref{fig:alignment}. Our method exhibits a steeper increase in IoU during the early stages compared to the spatial action map method~\cite{canberk2023cloth}. While both methods eventually converge to stable states, our approach achieves a higher final IoU value, indicating superior alignment performance.

Fig.~\ref{fig:simulatedalignment} displays the actions and the corresponding score maps. While the model structure is identical to that used in the smoothing task, the results exhibit distinct patterns observed in Fig.~\ref{fig:simulatedcoveragesteps}. Specifically, the key region of the score map is not limited to the edge of the garment, but includes the center areas, enabling large whole-body transformation during alignment. The goal of the action varies across different stages according to the state of the garment. For example, in Fig.~\ref{fig:simulatedalignment}, in the early stage (before Step 8), the actions mainly focus on smoothing the garment. Then, in the latter stages, the actions aim to rotate and transform the garment to align it with the predefined target goals. Notably, these actions (Step 5 to Step 20) may slightly fold the garment and create wrinkles, which demonstrates a shift in strategy as the process evolves.

\subsection{Ablation Study}
The specially proposed modules for garment manipulation tasks include the incorporation of the SEBlock, enlargement of the action area, fusion of background information, and the proposed shape loss for the alignment task. We perform systematic ablation experiments to evaluate module effectiveness and uncover their working principles. 

\subsubsection{Score map training}\label{sec:scoremaptraining}
First, the SEBlock is incorporated into the score head to re-weight the feature maps in channel-wise. Then, two methods are proposed to improve the score loss training, i.e., the action field enlargement and the background information fusion. These two methods address the imbalance between the foreground and background, each from a unique perspective: one by enlarging the action area and the other by incorporating background information.

\begin{table}[tbp]
\centering
\caption{Ablation Study Results for the Score Map Training}
\label{tab:score-ablation}
\begin{tabular}{cccccc}
\toprule
Virant & SEBlock & \makecell{Action\\ Enlarging} & \makecell{Background\\Information} & \makecell{Coverage} & \makecell{Relative\\Increase} \\
\midrule
 (a)  &            &            &            &0.812  &0.00 \\          
 (b)  &\checkmark  &            &            &0.817  &0.62 \\
 (c)  &            &\checkmark  &            &0.821  &1.11 \\
 (d)  &            &            &\checkmark  &0.816  &0.49 \\
 (e)  &\checkmark  &\checkmark  &            &0.844  &3.94 \\
 (f)  &\checkmark  &            &\checkmark  &0.825  &1.60 \\
 (g)  &            &\checkmark  &\checkmark  &0.845  &4.06 \\
 (h)  &\checkmark  &\checkmark  &\checkmark  &0.854  &5.17 \\
\bottomrule
\end{tabular}
\end{table}

We conduct ablation studies on the score head modules using the smoothing task, where the architecture preserves the complete scoring mechanism without interference from the supplementary components present in the alignment task. In Table~\ref{tab:score-ablation}, we present the results for the ``long sleeves'' clothing type, shown progressively. The variants range from the vanilla model to the full model that contains all the modules.

In Table~\ref{tab:score-ablation}, our vanilla model, which lacks any designed modules for score training, still achieves a higher coverage index than the state-of-the-art method reported by \cite{canberk2023cloth}. This result highlights the effectiveness of the overall dense action generator, demonstrating superior performance compared to the spatial action map method. This superiority may be attributed to the model's ability to continuously predict action parameters, i.e., angle and distance. 
In addition, in Table~\ref{tab:score-ablation}, the coverage index values show a progressive increase with sequential integration of the designed modules. Compared to the vanilla model, each of the three variants that incorporate individual modules independently improves the results, with relative increases ranging from 0.49\% to 1.11\%. These results demonstrate that the proposed modules can independently enhance the accuracy of score map prediction. The Action Enlarging module achieves the highest relative increase. Furthermore, with a combination of the two proposed modules, the accuracy improves further compared to variants with a single module, with increases between 1.60\% and 4.06\%, indicating that the modules do not interfere with each other. Ultimately, the complete model achieves the highest accuracy, marking a significant relative increase of 5.17\% over the vanilla model. 
\begin{figure}[tbp]
    \centering
    \includegraphics[width=0.9\columnwidth]{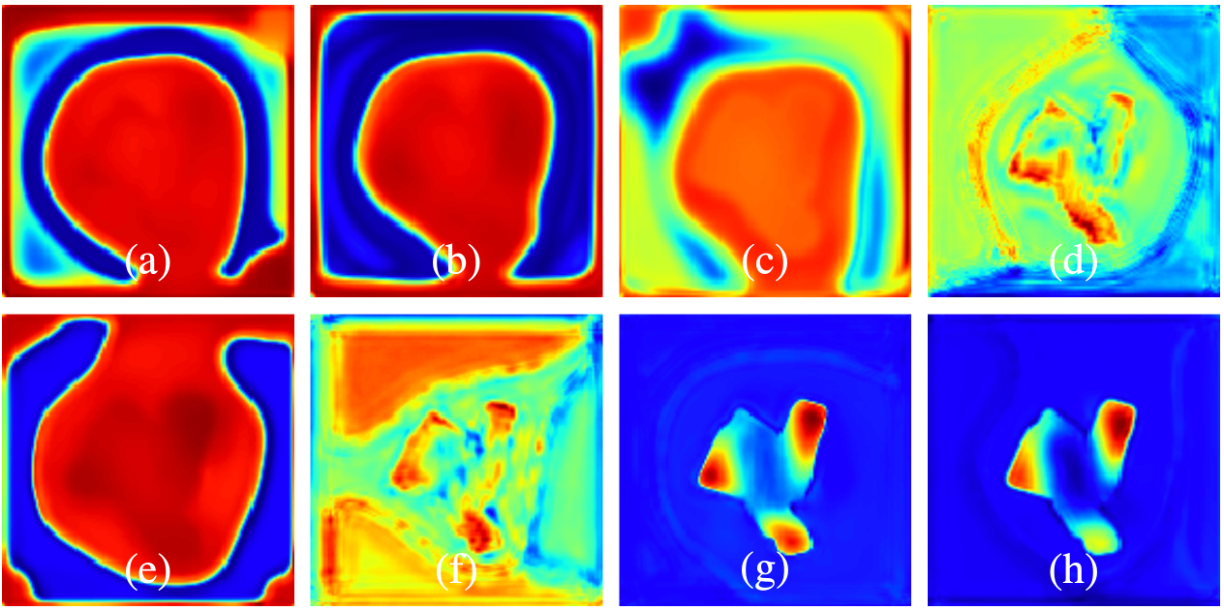}
    \caption{Score map comparison of each model variant in ablation study.}
    \label{fig:ablation-score}
\end{figure}

Furthermore, Fig.~\ref{fig:ablation-score} presents the score maps for each of the models in the ablation experiment. Compared with images (d) and (f)-(h), which have background information integration, images (a)-(c) and (e) erroneously highlight the background, especially at the edge and corner areas, and fail to clearly output the contour of the garment. This suggests a potential overfitting issue due to an information imbalance between foreground and background. Then, comparing image (d) to images (g) and (h), it shows that only the background information module still struggles to capture high-accuracy semantic details. However, the addition of an Action Enlarging module significantly reduces noise and clarifies the garment contours, indicating that the combination of these two modules effectively learns semantic information. Notably, the model automatically and precisely retrieves the manipulation areas, i.e., corner areas in the case of the garment smoothing task, which aligns with common practices of smoothing a garment with single-point interaction. Finally, when comparing images (g) and (h), the SEBlock module appears to guide the model to focus on the most important areas, further refining the score map. Overall, the ablation experiment results of score map training show that all proposed modules contribute to enhancing the accuracy of the score. 

\begin{figure}[tbp]
    \centering
    \includegraphics[width=0.9\columnwidth]{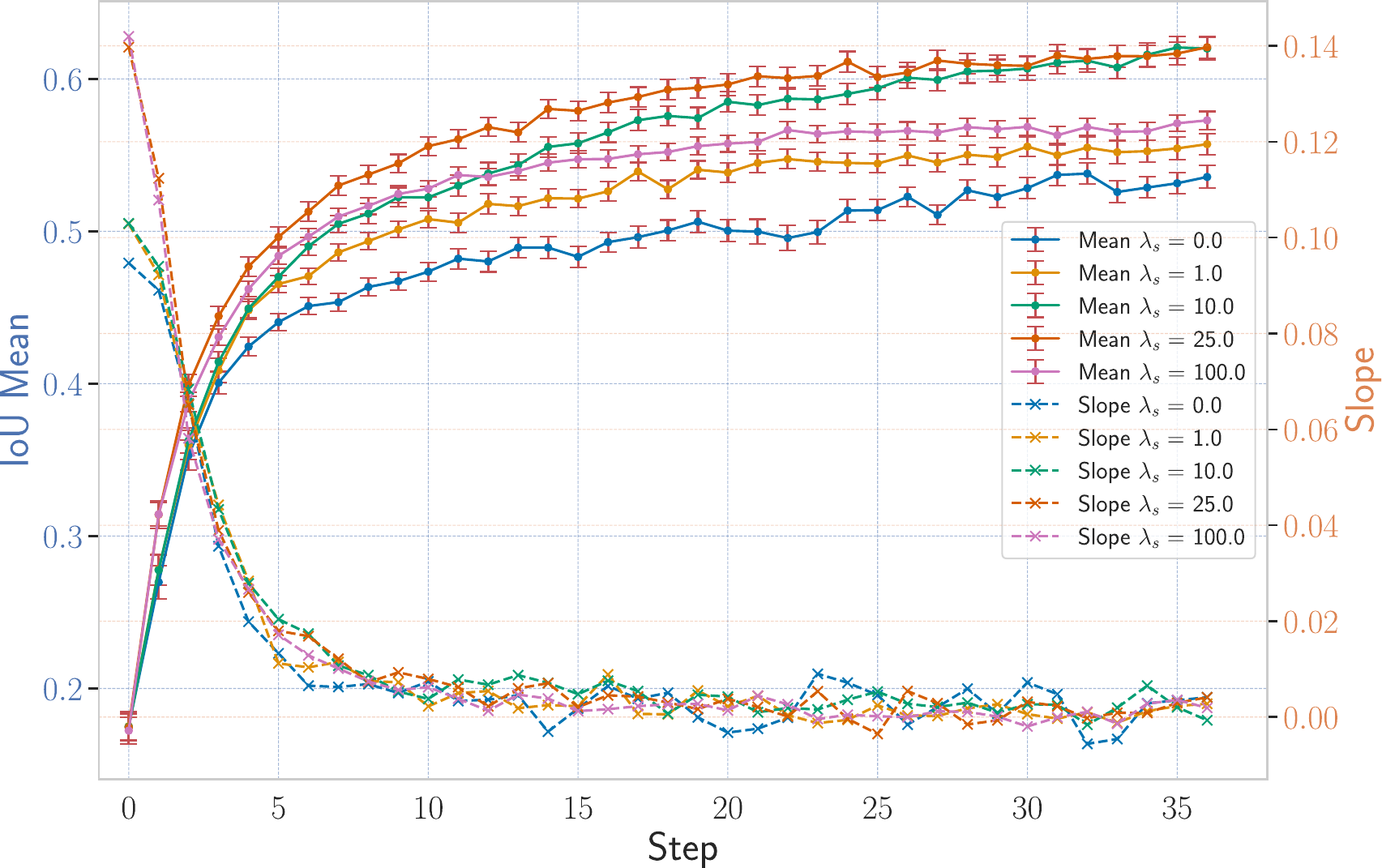}
    \caption{IoU values of the alignment task of the model trained with the shape loss.}
    \label{fig:ablation-shape-iou}
\end{figure}

\subsubsection{Shape loss}\label{sec:ablationshapeloss}
The shape loss learns an action field that moves the pixels of the garment to the positions specified by the target state. We evaluate its effect by training the action generator with different shape loss contributions, achieved through adjusting the shape loss weight $\lambda_s$. 

As shown in Fig.~\ref{fig:ablation-shape-iou}, the IoU values of the model with effective shape loss, i.e., with $\lambda_s \in [1.0, 10.0, 25.0, 100.0]$, are larger than that of the baseline model with no contributions from the shape loss, i.e., with a $\lambda_s = 0$, in most steps and ultimately stabilize, demonstrating the effectiveness of adopting shape loss. This also confirms the feasibility of simulating actions of the action field through pixel movements. Furthermore, the slopes of the IoU curves for the models with shape loss are larger than those of the vanilla model in the early stages, indicating that it is more effective when the garments' states are at greater discrepancies to the target and thus in a more challenging state. This might be because the shape loss model is also supervised by the target state, which is less influenced by the distances of the labeled actions. In contrast, the action field learned through the shape loss does not limit the scale of action, allowing for larger adjustments that quickly align the garment in the early stages. 

\begin{table}[tbp]
    \centering
    \caption{Ablation Results on the Alignment Task}
    \label{tab:ablation-shape}
    \begin{tabular}{c c cc}
    \toprule
    \multirow{2}{*}{\textbf{Variants}} 
    & \multirow{2}{*}{\textbf{Shape Weights}} 
    & \multicolumn{2}{c}{\textbf{IoU}} \\
    \cmidrule(lr){3-4}
    &  & \textbf{Step10} & \textbf{Step36} \\
    \midrule
    (1) & 0.0    & 0.482  & 0.536 \\ 
    (2) & 1.0    & 0.506  & 0.557 \\
    (3) & 10.0   & 0.530  & 0.612 \\
    (4) & 25.0   & 0.560  & 0.621 \\
    (5) & 100.0  & 0.537  & 0.573 \\
    \bottomrule
    \end{tabular}
\end{table}

To further investigate the proposed shape loss, we present the detailed evaluation results in Table~\ref{tab:ablation-shape}. The baseline model with $\lambda_{s} = 0.0$ records an IoU value of 0.482 after 10 steps of planar action, and it finally reaches 0.536 after 36 steps, illustrating the effectiveness of planar action in garment alignment. As the shape loss appears and its contribution gains with the $\lambda_{s}$ increasing to 1.0, 10.0, and 25.0, the variants achieve higher IoU values in both Step10 and Step36, simultaneously and progressively. However, once the shape weight $\lambda_{s}$ reaches 100.0, the IoU values of Step10 and Step36 fall back to 0.537 and 0.573. This suggests that an overemphasis on shape similarity may not optimally align with other critical factors, such as overall pose error. Thus, the contribution of sampled labels should also be balanced to achieve the highest IoU value. 

While the shape loss yields higher IoU values, it focuses on image-level similarity and ignores garment details, potentially leading to local minima. For example, the garment may result in an upside-down position. In contrast, sampled actions manipulate the garment to gradually reduce the average point-to-point distance to the target, but lack the knowledge of the final pose. Thus, balancing the contributions of shape loss and sampled action errors through a combined loss function, calculated as Equation (\ref{eq:total_loss}), is crucial for achieving optimal garment alignment. Note that for the case when $\lambda_s = 100.0$, the shape loss contributes much more than the other regression losses, $L_{angle}$ and $L_{distance}$, dominating the training of the model. This dominance suggests that the shape loss module can achieve the alignment task by itself, functioning as an unsupervised regression, which does not require other labels, i.e., $gt_{sin}$, $gt_{cos}$, and $gt_{l}$, to train the model. 

\begin{figure}[tbp]
\centering
    \includegraphics[width=0.9\columnwidth]{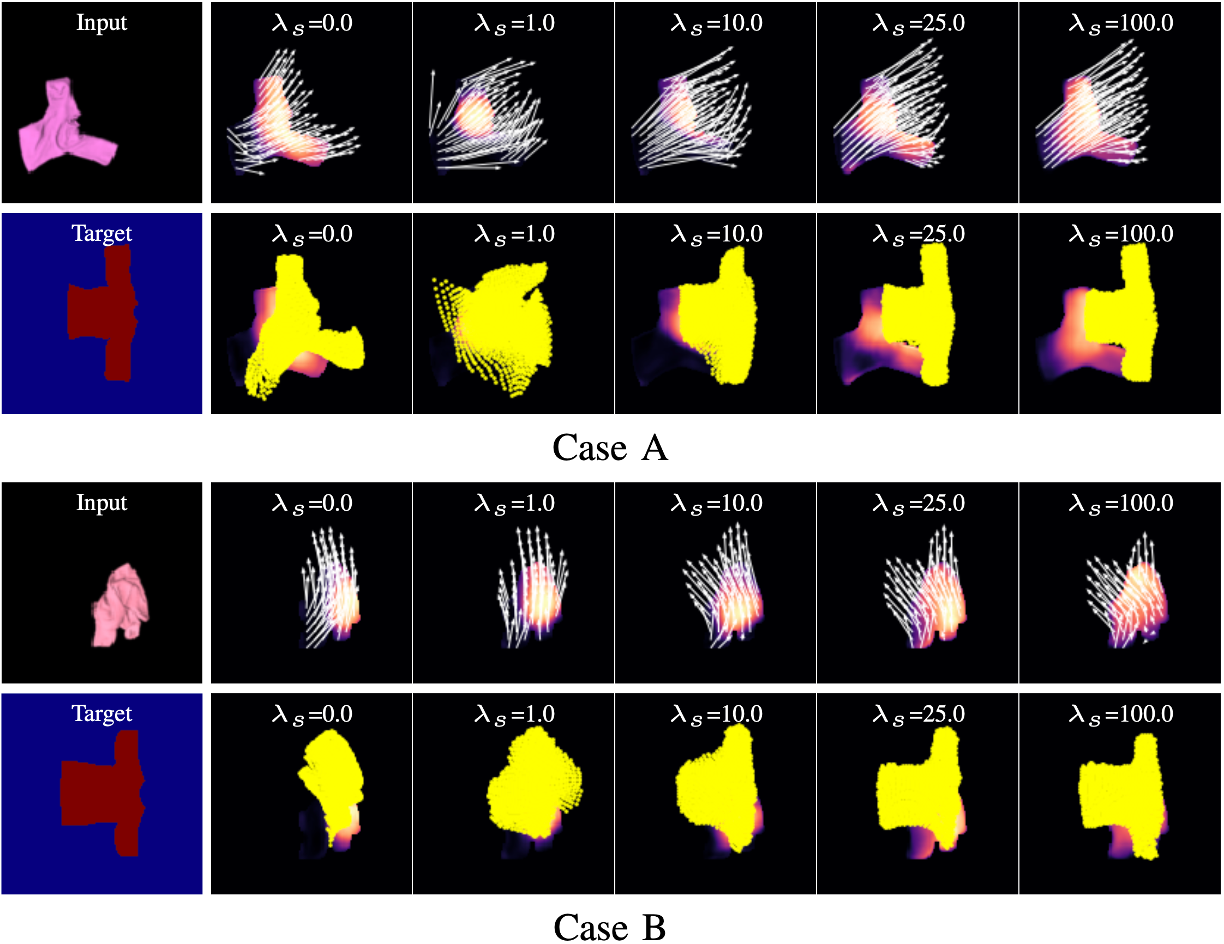}
    \caption{Visualizing the impact of shape loss. Case A and Case B correspond to different initial states. For both cases, the top row displays the input image and the generated dense actions (white arrows) under different shape loss weights, while the bottom row displays target masks and the results of pixel movements (yellow dots) by these actions.}
\label{fig:ablation-shape-loss-positive}
\end{figure}

As described in Section~\ref{sec:shapeloss}, the movement of a single pixel determined by each of the dense actions can approximate the overall manipulation result. We provide visualization of actions and the image results in Fig.~\ref{fig:ablation-shape-loss-positive}. As shown in Case A of Fig.~\ref{fig:ablation-shape-loss-positive}, when $\lambda_s$ increases from $0.0$ to $100.0$, the model is more heavily supervised by the shape loss, making the resulting shape closer to the target mask. Specifically, without the shape loss ($\lambda_s=0.0$), most of the actions have similar angles, which only have the effect of transforming the garment area. Then, as the weight $\lambda_s$ increases from 0.0 to 1.0, the shape loss begins to occur and exerts its effect. Consequently, the overall result is that the dense pixel movements lead to a smoothing effect, which enlarges the size of the garment area. Finally, as the weight increases to a value such as $\lambda_s \in[10.0, 25.0, 100.0]$, the contours of the result areas are clearer and more similar to the target mask, compared with the results with $\lambda_s=[0.0, 1.0]$. This indicates that the model can generate actions that effectively manipulate the garment to align with the target. This observation is consistent with the quantitative results presented in Table~\ref{tab:ablation-shape} and elucidates the decline in the IoU when $\lambda_s$ is excessively increased. 

On the other hand, for Case B shown in Fig.~\ref{fig:ablation-shape-loss-positive}, the trends in pixel movements with changes in $\lambda_s$ are similar to the positive case, despite the differences in the initial pose compared to Case A. This suggests that our method employs a consistent strategy regardless of initial conditions. In summary, the shape loss itself focuses on image-level guidance by reducing the shape difference between the result and the target. This implies that a model trained only with shape loss fails to retrieve the inner structure and semantic information from the garment image, and struggles with extreme poses. For this reason, shape loss requires the integration with the supervised learning of the angle and distance.

\begin{figure}[tbp]
    \centering
    \includegraphics[width=0.9\columnwidth]{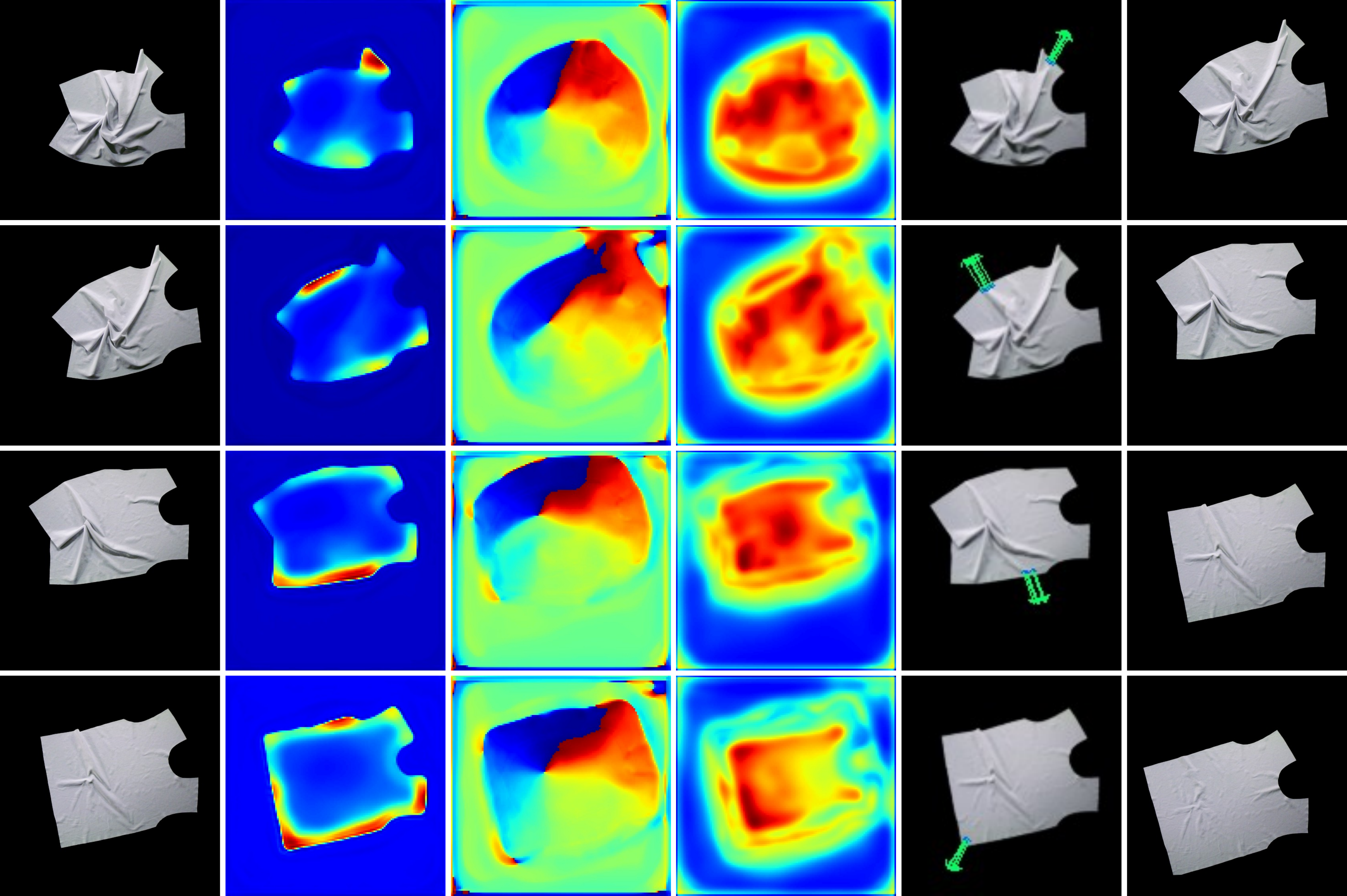}
    \caption{Steps of a T-shirt-shaped fabric smoothing trial. From left to right, the images depict the initial state, score map, angle map, distance map, visualization of the action, and final state (which serves as the initial state for the next step). From top to bottom, each row represents a single step.}
    \label{fig:realexperiment}
\end{figure}

\begin{figure*}[tbp]
    \centering 
    \begin{subfigure}{0.9\columnwidth}
        \includegraphics[width=\textwidth]{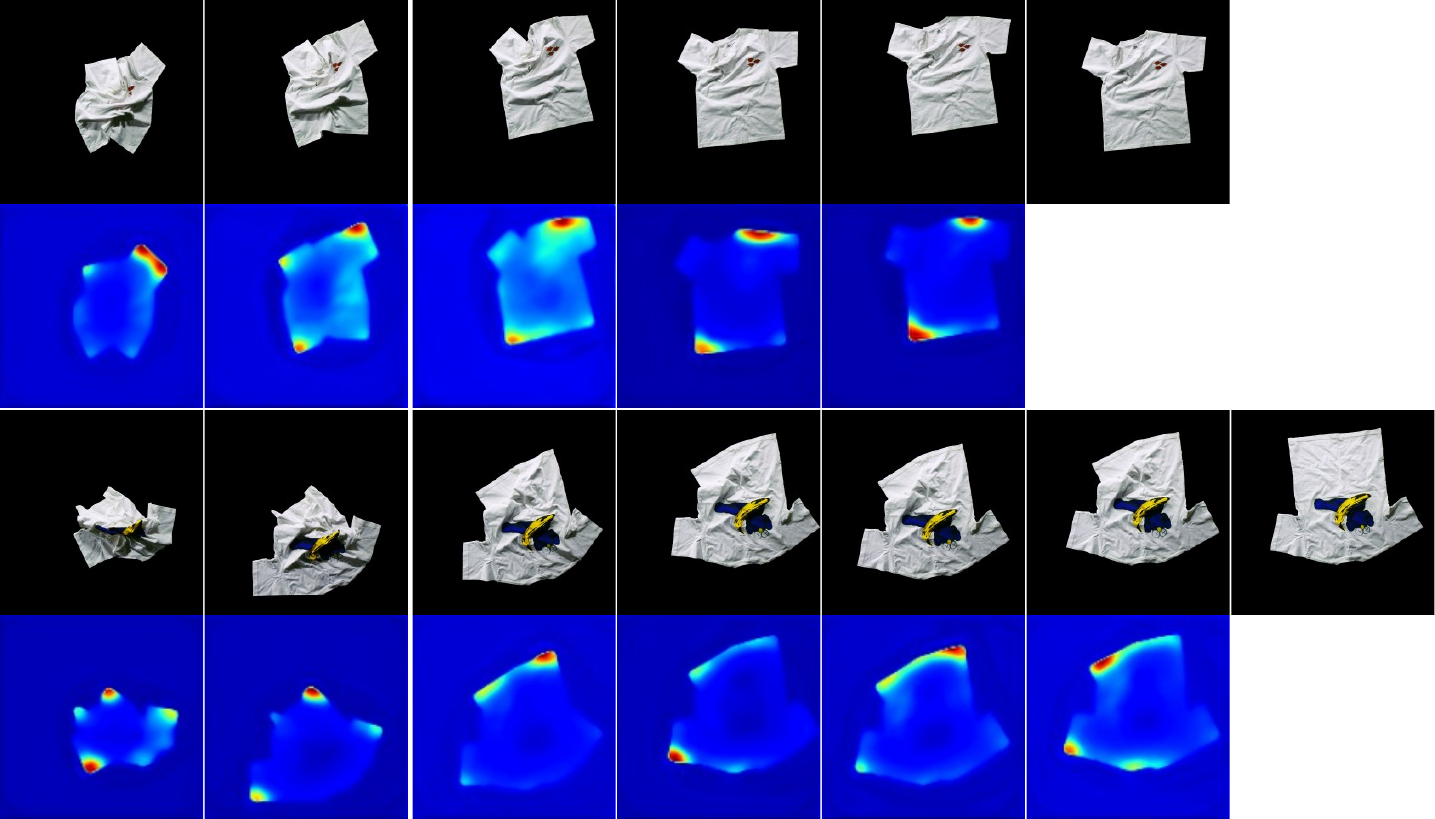}
        \caption{T-shirt}
        \label{fig:tshirts}
    \end{subfigure} 
    \begin{subfigure}{0.9\columnwidth}
        \includegraphics[width=\textwidth]{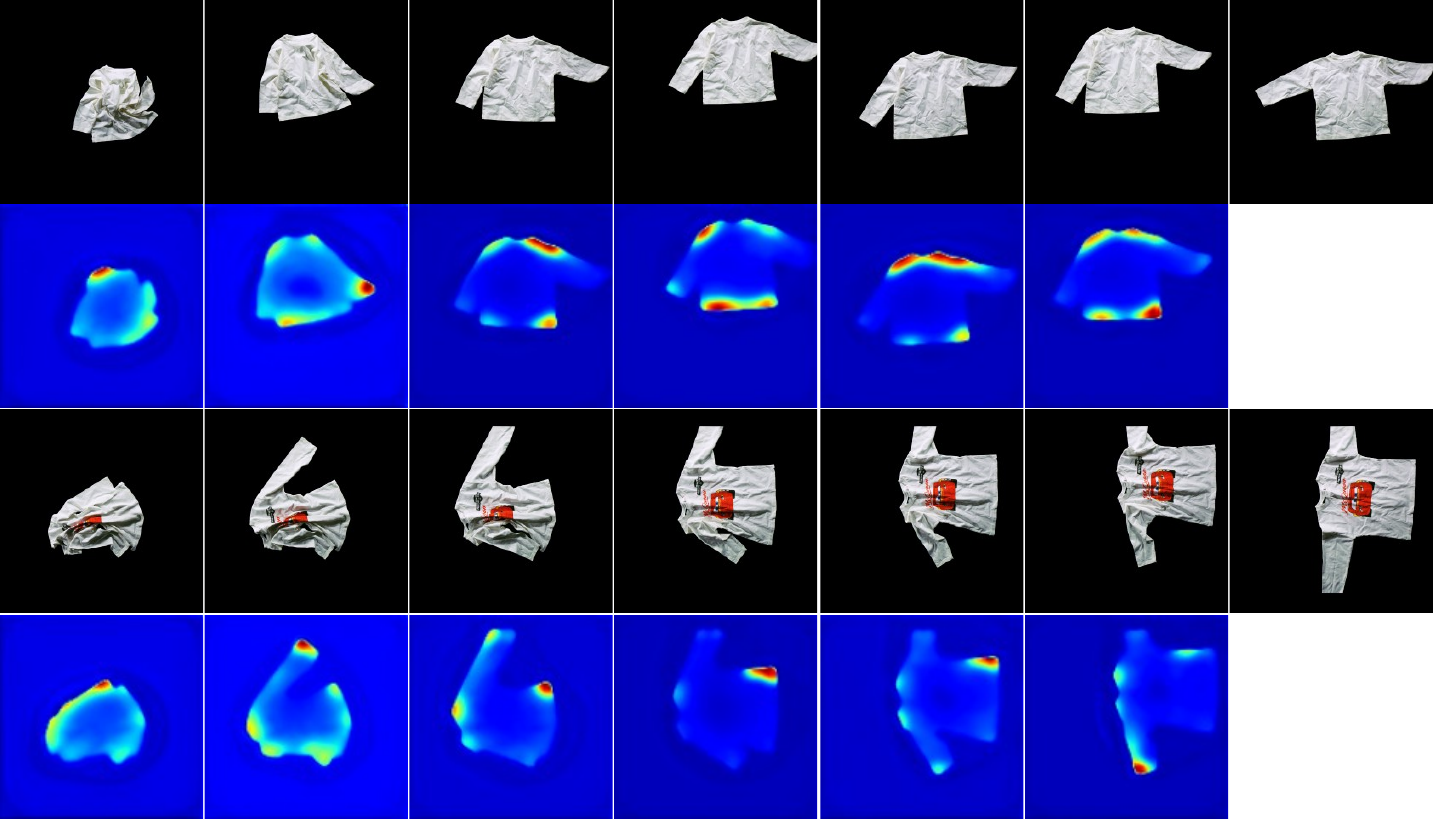}
        \caption{Long sleeves}
        \label{fig:longsleeve}
    \end{subfigure} \\
    \caption{Selected results of the garment smoothing experiment of the T-shirts and long sleeves. Each trial would be terminated early if the coverage index value stabilizes or reaches a threshold. }
    \label{fig:morerealexperiments}
\end{figure*}

\subsection{Real-world Experiments}\label{sec:realworldexperiments}
To further evaluate the effectiveness of the proposed robot garment manipulation system and assess the simulation-to-real gap, we conduct real-world experiments using our network trained only on simulated data. We tested the system on seven different upper garments, including vests, T-shirts, and long sleeves. Each garment contains 5 trials in which the garment is manipulated from randomly crumpled states by the robot. A trial is stopped when the coverage index is stable in consecutive actions or reaches a maximum of 10 actions. The results are present in Table~\ref{tab:realworldunfolding}. It can be observed that the models trained solely by the synthetic garment images in the simulator demonstrate high coverage indexes in the real-world experiment. Compared with Cloth Funnels \cite{canberk2023cloth}, the robot with our action generator achieves higher coverage index values after the first 3 actions (Step3) and when the trial is stopped (StepFinal), further demonstrating its superior performance.

\begin{table}[tbp]
    \centering
    \caption{Real-world Garment Smoothing Experiment}
    \label{tab:realworldunfolding}
    \begin{tabular}{ccc}
    \toprule
    \multirow{2}{*}{\textbf{Methods}} 
    & \multicolumn{2}{c}{\textbf{Coverage}} \\
    \cmidrule(lr){2-3}
    & \textbf{Step3} & \textbf{StepFinal} \\
    \midrule
    Cloth Funnels \cite{canberk2023cloth} & 0.814  & 0.833 \\ 
    Ours & 0.851  & 0.865 \\
    \bottomrule
    \end{tabular}
\end{table}

Fig.~\ref{fig:realexperiment} presents an example of a real-world smoothing sequence, with each row showing the fabric state after executing one pulling action. The model takes the fabric image as input, produces the score, angle, and distance map to reconstruct the dense actions. The actions with the top 5 scores are visualized, and the robot executes the highest scoring one. This procedure repeats, using the resulting state as the new input, until the trial reaches the stopping condition. As shown in Fig.~\ref{fig:realexperiment}, our robotic system progressively smooths crumpled fabric, yielding less wrinkled fabric at each step. Ultimately, the fabric is completely flattened, demonstrating our system's potential for real-world deployment with minimal adaptation. 

Fig.~\ref{fig:morerealexperiments} shows examples of the experimental results using different garment types, including T-shirts and long sleeves. Our model achieves good smoothing performance on the novel T-shirts and long sleeves. The score maps demonstrate that the model can consistently highlight appropriate pulling start regions for manipulation and adjust the score distribution according to the current garment configuration. \href{https://garmentaction.pages.dev/}{We also provide supplementary videos to demonstrate the experiments.}

\section{Conclusion and Future Work}
In this paper, we propose a robotic system capable of automatically manipulating various categories of garments. The vision-based generation model is specifically designed for the single-point planar action. We introduce a novel framework to generate dense actions by single model forward propagation, significantly reducing the computation time while ensuring the prediction of continuous action parameters. For action score generation, we identify class imbalance as the primary cause of performance degradation and suggest incorporating the background semantic information to address this issue. Furthermore, we leverage target masks and shape similarity metrics to guide the model training, enhancing manipulation accuracy without additional computational overhead. Extensive experiments in both simulated and real-world environments demonstrate the superior performance of our method. 

Currently, our shape loss strategy shows the potential to directly guide generation actions that move the garment to a predefined target configuration. In the future, we plan to use the target information as the conditional input to the model, thereby enabling it to align the garment to an arbitrarily specified rotation and position. In addition, we plan to extend our shape similarity supervision to 3D scenes, where both actions and garments representation are expressed in spatial formats, i.e., spatial vector and point clouds. In this setting, potential deformation of dense actions can be generated on the point clouds through spatial forward wrapping. This adaptation of the shape loss module to the spatial domain is expected to enhance the accuracy of geometric modeling and improve the overall robustness of the framework. 

\bibliographystyle{IEEEtran}
\bibliography{references}

@inproceedings{he2016deep,
  title={Deep residual learning for image recognition},
  author={He, Kaiming and Zhang, Xiangyu and Ren, Shaoqing and Sun, Jian},
  booktitle={Proceedings of the IEEE conference on computer vision and pattern recognition},
  pages={770--778},
  year={2016}
}

@article{cheng2022robot,
  title={A Robot Grasping System With Single-Stage Anchor-Free Deep Grasp Detector},
  author={Cheng, Hu and Wang, Yingying and Meng, Max Q-H},
  journal={IEEE Transactions on Instrumentation and Measurement},
  volume={71},
  pages={1--12},
  year={2022},
  publisher={IEEE}
}

@inproceedings{canberk2023cloth,
  title={Cloth funnels: Canonicalized-alignment for multi-purpose garment manipulation},
  author={Canberk, Alper and Chi, Cheng and Ha, Huy and Burchfiel, Benjamin and Cousineau, Eric and Feng, Siyuan and Song, Shuran},
  booktitle={2023 IEEE International Conference on Robotics and Automation (ICRA)},
  pages={5872--5879},
  year={2023},
  organization={IEEE}
}

@inproceedings{ha2022flingbot,
  title={Flingbot: The unreasonable effectiveness of dynamic manipulation for cloth unfolding},
  author={Ha, Huy and Song, Shuran},
  booktitle={Conference on Robot Learning},
  pages={24--33},
  year={2022},
  organization={PMLR}
}

@inproceedings{lin2021softgym,
  title={Softgym: Benchmarking deep reinforcement learning for deformable object manipulation},
  author={Lin, Xingyu and Wang, Yufei and Olkin, Jake and Held, David},
  booktitle={Conference on Robot Learning},
  pages={432--448},
  year={2021},
  organization={PMLR}
}

@inproceedings{weng2022fabricflownet,
  title={Fabricflownet: Bimanual cloth manipulation with a flow-based policy},
  author={Weng, Thomas and Bajracharya, Sujay Man and Wang, Yufei and Agrawal, Khush and Held, David},
  booktitle={Conference on Robot Learning},
  pages={192--202},
  year={2022},
  organization={PMLR}
}

@inproceedings{agarwal2023point,
  title={Point-based Correspondence Estimation for Cloth Alignment and Manipulation},
  author={Agarwal, Mansi and Weng, Thomas and Held, David},
  booktitle={RSS 2023 Workshop on Symmetries in Robot Learning}
}

@article{luo2024multi,
  title={Multi-stage cable routing through hierarchical imitation learning},
  author={Luo, Jianlan and Xu, Charles and Geng, Xinyang and Feng, Gilbert and Fang, Kuan and Tan, Liam and Schaal, Stefan and Levine, Sergey},
  journal={IEEE Transactions on Robotics},
  year={2024},
  publisher={IEEE}
}

@inproceedings{chen2023autobag,
  title={Autobag: Learning to open plastic bags and insert objects},
  author={Chen, Lawrence Yunliang and Shi, Baiyu and Seita, Daniel and Cheng, Richard and Kollar, Thomas and Held, David and Goldberg, Ken},
  booktitle={2023 IEEE International Conference on Robotics and Automation (ICRA)},
  pages={3918--3925},
  year={2023},
  organization={IEEE}
}

@article{wu2020spatial,
  title={Spatial action maps for mobile manipulation},
  author={Wu, Jimmy and Sun, Xingyuan and Zeng, Andy and Song, Shuran and Lee, Johnny and Rusinkiewicz, Szymon and Funkhouser, Thomas},
  journal={arXiv preprint arXiv:2004.09141},
  year={2020}
}

@inproceedings{zeng2018learning,
  title={Learning synergies between pushing and grasping with self-supervised deep reinforcement learning},
  author={Zeng, Andy and Song, Shuran and Welker, Stefan and Lee, Johnny and Rodriguez, Alberto and Funkhouser, Thomas},
  booktitle={2018 IEEE/RSJ International Conference on Intelligent Robots and Systems (IROS)},
  pages={4238--4245},
  year={2018},
  organization={IEEE}
}

@inproceedings{wu2024unigarmentmanip,
  title={UniGarmentManip: A Unified Framework for Category-Level Garment Manipulation via Dense Visual Correspondence},
  author={Wu, Ruihai and Lu, Haoran and Wang, Yiyan and Wang, Yubo and Dong, Hao},
  booktitle={Proceedings of the IEEE/CVF Conference on Computer Vision and Pattern Recognition},
  pages={16340--16350},
  year={2024}
}

@article{schmidt2016self,
  title={Self-supervised visual descriptor learning for dense correspondence},
  author={Schmidt, Tanner and Newcombe, Richard and Fox, Dieter},
  journal={IEEE Robotics and Automation Letters},
  volume={2},
  number={2},
  pages={420--427},
  year={2016},
  publisher={IEEE}
}

@inproceedings{gao2023iterative,
  title={Iterative interactive modeling for knotting plastic bags},
  author={Gao, Chongkai and Li, Zekun and Gao, Haichuan and Chen, Feng},
  booktitle={Conference on Robot Learning},
  pages={571--582},
  year={2023},
  organization={PMLR}
}

@inproceedings{sundaresan2020learning,
  title={Learning rope manipulation policies using dense object descriptors trained on synthetic depth data},
  author={Sundaresan, Priya and Grannen, Jennifer and Thananjeyan, Brijen and Balakrishna, Ashwin and Laskey, Michael and Stone, Kevin and Gonzalez, Joseph E and Goldberg, Ken},
  booktitle={2020 IEEE International Conference on Robotics and Automation (ICRA)},
  pages={9411--9418},
  year={2020},
  organization={IEEE}
}

@inproceedings{ganapathi2021learning,
  title={Learning dense visual correspondences in simulation to smooth and fold real fabrics},
  author={Ganapathi, Aditya and Sundaresan, Priya and Thananjeyan, Brijen and Balakrishna, Ashwin and Seita, Daniel and Grannen, Jennifer and Hwang, Minho and Hoque, Ryan and Gonzalez, Joseph E and Jamali, Nawid and others},
  booktitle={2021 IEEE International Conference on Robotics and Automation (ICRA)},
  pages={11515--11522},
  year={2021},
  organization={IEEE}
}

@article{achiam2023gpt,
  title={Gpt-4 technical report},
  author={Achiam, Josh and Adler, Steven and Agarwal, Sandhini and Ahmad, Lama and Akkaya, Ilge and Aleman, Florencia Leoni and Almeida, Diogo and Altenschmidt, Janko and Altman, Sam and Anadkat, Shyamal and others},
  journal={arXiv preprint arXiv:2303.08774},
  year={2023}
}

@inproceedings{clark2023household,
  title={Household clothing set and benchmarks for characterising end-effector cloth manipulation},
  author={Clark, Angus B and Cramphorn-Neal, Luke and Rachowiecki, Michal and Gregg-Smith, Austin},
  booktitle={2023 IEEE International Conference on Robotics and Automation (ICRA)},
  pages={9211--9217},
  year={2023},
  organization={IEEE}
}

@inproceedings{chen2023learning,
  title={Learning to grasp clothing structural regions for garment manipulation tasks},
  author={Chen, Wei and Lee, Dongmyoung and Chappell, Digby and Rojas, Nicolas},
  booktitle={2023 IEEE/RSJ International Conference on Intelligent Robots and Systems (IROS)},
  pages={4889--4895},
  year={2023},
  organization={IEEE}
}

@inproceedings{wang2024trtm,
  title={Trtm: Template-based reconstruction and target-oriented manipulation of crumpled cloths},
  author={Wang, Wenbo and Li, Gen and Zamora, Miguel and Coros, Stelian},
  booktitle={2024 IEEE International Conference on Robotics and Automation (ICRA)},
  pages={12522--12528},
  year={2024},
  organization={IEEE}
}

@inproceedings{lin2022learning,
  title={Learning visible connectivity dynamics for cloth smoothing},
  author={Lin, Xingyu and Wang, Yufei and Huang, Zixuan and Held, David},
  booktitle={Conference on Robot Learning},
  pages={256--266},
  year={2022},
  organization={PMLR}
}

@inproceedings{chi2021garmentnets,
  title={Garmentnets: Category-level pose estimation for garments via canonical space shape completion},
  author={Chi, Cheng and Song, Shuran},
  booktitle={Proceedings of the IEEE/CVF International Conference on Computer Vision},
  pages={3324--3333},
  year={2021}
}

@inproceedings{blanco2023qdp,
  title={QDP: Learning to sequentially optimise quasi-static and dynamic manipulation primitives for robotic cloth manipulation},
  author={Blanco-Mulero, David and Alcan, Gokhan and Abu-Dakka, Fares J and Kyrki, Ville},
  booktitle={2023 IEEE/RSJ International Conference on Intelligent Robots and Systems (IROS)},
  pages={984--991},
  year={2023},
  organization={IEEE}
}

@inproceedings{hu2018squeeze,
  title={Squeeze-and-excitation networks},
  author={Hu, Jie and Shen, Li and Sun, Gang},
  booktitle={Proceedings of the IEEE conference on computer vision and pattern recognition},
  pages={7132--7141},
  year={2018}
}

@inproceedings{berenson2013manipulation,
  title={Manipulation of deformable objects without modeling and simulating deformation},
  author={Berenson, Dmitry},
  booktitle={2013 IEEE/RSJ International Conference on Intelligent Robots and Systems},
  pages={4525--4532},
  year={2013},
  organization={IEEE}
}

@inproceedings{matas2018sim,
  title={Sim-to-real reinforcement learning for deformable object manipulation},
  author={Matas, Jan and James, Stephen and Davison, Andrew J},
  booktitle={Conference on Robot Learning},
  pages={734--743},
  year={2018},
  organization={PMLR}
}

@ARTICLE{zhou2025dualarm,
  author={Zhou, Changshi and Jiang, Rong and Luan, Feng and Meng, Shaoqiang and Wang, Zhipeng and Dong, Yanchao and Zhou, Yanmin and He, Bin},
  journal={IEEE/ASME Transactions on Mechatronics}, 
  title={Dual-Arm Robotic Fabric Manipulation With Quasi-Static and Dynamic Primitives for Rapid Garment Flattening}, 
  year={2025},
  volume={},
  number={},
  pages={1-11},
  doi={10.1109/TMECH.2025.3556283}}

@inproceedings{yang2024clothppo,
  title={ClothPPO: a proximal policy optimization enhancing framework for robotic cloth manipulation with observation-aligned action spaces},
  author={Yang, Libing and Li, Yang and Chen, Long},
  booktitle={Proceedings of the Thirty-Third International Joint Conference on Artificial Intelligence},
  pages={6895--6903},
  year={2024}
}

@inproceedings{florence2018dense,
  title={Dense Object Nets: Learning Dense Visual Object Descriptors By and For Robotic Manipulation},
  author={Florence, Peter R and Manuelli, Lucas and Tedrake, Russ},
  booktitle={Conference on Robot Learning},
  pages={373--385},
  year={2018},
  organization={PMLR}
}

@inproceedings{wu2019learning,
  title={Learning to Manipulate Deformable Objects without Demonstrations},
  author={Wu, Yilin and Yan, Wilson and Kurutach, Thanard and Pinto, Lerrel and Abbeel, Pieter},
  booktitle={Robotics: Science and Systems},
  year={2020}
}

@inproceedings{huang2022mesh,
    AUTHOR    = {Zixuan Huang AND Xingyu Lin AND David Held}, 
    TITLE     = {{Mesh-based Dynamics with Occlusion Reasoning for Cloth Manipulation}}, 
    BOOKTITLE = {Proceedings of Robotics: Science and Systems}, 
    YEAR      = {2022}, 
    ADDRESS   = {New York City, NY, USA}, 
    MONTH     = {June}, 
    DOI       = {10.15607/RSS.2022.XVIII.011} 
}

@inproceedings{li2018learning,
  title={Learning Particle Dynamics for Manipulating Rigid Bodies, Deformable Objects, and Fluids},
  author={Li, Yunzhu and Wu, Jiajun and Tedrake, Russ and Tenenbaum, Joshua B and Torralba, Antonio},
  booktitle={International Conference on Learning Representations},
  year={2019}
}

@ARTICLE{10132389,
  author={Cheng, Hu and Wang, Yingying and Meng, Max Q. -H.},
  journal={IEEE Transactions on Automation Science and Engineering}, 
  title={Anchor-Based Multi-Scale Deep Grasp Pose Detector With Encoded Angle Regression}, 
  year={2024},
  volume={21},
  number={3},
  pages={3130-3142}
}

@article{li2018model,
  title={Model-driven feedforward prediction for manipulation of deformable objects},
  author={Li, Yinxiao and Wang, Yan and Yue, Yonghao and Xu, Danfei and Case, Michael and Chang, Shih-Fu and Grinspun, Eitan and Allen, Peter K},
  journal={IEEE Transactions on Automation Science and Engineering},
  volume={15},
  number={4},
  pages={1621--1638},
  year={2018},
  publisher={IEEE}
}

@ARTICLE{10966003,
  author={Zhou, Changshi and Xu, Haichuan and Hu, Jiarui and Luan, Feng and Wang, Zhipeng and Dong, Yanchao and Zhou, Yanmin and He, Bin},
  journal={IEEE Transactions on Automation Science and Engineering}, 
  title={SSFold: Learning to Fold Arbitrary Crumpled Cloth Using Graph Dynamics From Human Demonstration}, 
  year={2025},
  volume={22},
  number={},
  pages={14448-14460}}

@article{cheng2026vision,
  title={Vision-based single-stage grasp pose estimator with rotated anchors and automatic label generation},
  author={Cheng, Hu and Wang, Yingying and Meng, Max Q-H},
  journal={Science China Information Sciences},
  volume={69},
  number={1},
  pages={112204},
  year={2026},
  publisher={Springer}
}

@article{tokuda2025transformer,
  title={Transformer Driven Visual Servoing for Fabric Texture Matching Using Dual-Arm Manipulator},
  author={Tokuda, Fuyuki and Seino, Akira and Kobayashi, Akinari and Tang, Kai and Kosuge, Kazuhiro},
  journal={IEEE Robotics and Automation Letters},
  volume={11},
  number={2},
  pages={1522--1529},
  year={2025},
  publisher={IEEE}
}

@article{huang2025sis,
  title={SIS: Seam-Informed Strategy for T-Shirt Unfolding},
  author={Huang, Xuzhao and Seino, Akira and Tokuda, Fuyuki and Kobayashi, Akinari and Chen, Dayuan and Hirata, Yasuhisa and Tien, Norman C and Kosuge, Kazuhiro},
  journal={IEEE Robotics and Automation Letters},
  year={2025},
  publisher={IEEE}
}

@inproceedings{deng2025general,
  title={General-purpose clothes manipulation with semantic keypoints},
  author={Deng, Yuhong and Hsu, David},
  booktitle={2025 IEEE International Conference on Robotics and Automation (ICRA)},
  pages={13181--13187},
  year={2025},
  organization={IEEE}
}

@inproceedings{GPTFabric2024,
    title   = {GPT-Fabric: Smoothing and Folding Fabric by Leveraging Pre-Trained Foundation Models},
    author  = {Vedant Raval and Enyu Zhao and Hejia Zhang and Stefanos Nikolaidis and Daniel Seita},
    booktitle = {The International Symposium of Robotics Research (ISRR)},
    Year    = {2024}
}

@INPROCEEDINGS{Xu-RSS-22, 
    AUTHOR    = {Zhenjia Xu AND Cheng Chi AND Benjamin Burchfiel AND Eric Cousineau AND Siyuan Feng AND Shuran Song}, 
    TITLE     = {{DextAIRity: Deformable Manipulation Can be a Breeze}}, 
    BOOKTITLE = {Proceedings of Robotics: Science and Systems}, 
    YEAR      = {2022}, 
    ADDRESS   = {New York City, NY, USA}, 
    MONTH     = {June}, 
    DOI       = {10.15607/RSS.2022.XVIII.017} 
}

\end{document}